\documentclass{article}
\usepackage{iclr2027_conference,times}
\iclrpreprintcopy

\usepackage[utf8]{inputenc}
\usepackage[T1]{fontenc}
\usepackage{hyperref}
\hypersetup{pdftitle={PrefMoE: Mitigating Preference Collapse in Personalized Multimodal Large Language Models},pdfauthor={Fan Lyu; Wenqi Zhang; Joost van de Weijer},pdfsubject={Preprint},pdfkeywords={personalized multimodal large language models, preference collapse, mixture of experts}}
\usepackage{url}
\usepackage{booktabs}
\usepackage{amsfonts}
\usepackage{nicefrac}
\usepackage{microtype}
\usepackage{xcolor}
\usepackage{makecell}
\usepackage{graphicx}
\usepackage{amssymb}
\usepackage{float}
\usepackage{amsmath}
\usepackage{multirow}
\usepackage{enumerate,enumitem}
\usepackage{pifont}
\usepackage{wrapfig}
\usepackage{caption}
\usepackage{placeins}  
\usepackage{needspace}

\newcommand{\cmark}{\ding{51}}

\definecolor{burgundy}{RGB}{139,0,35}
\definecolor{revma}{RGB}{255,255,255}
\def\lyu{\textcolor{black}}
\def\zwq{\textcolor{black}}

\title{PrefMoE: Mitigating Preference Collapse in Personalized Multimodal Large Language Models}
\author{Fan Lyu, Wenqi Zhang, and Joost van de Weijer\\
\normalfont
Computer Vision Center, Universitat Autònoma de Barcelona
Barcelona, Spain, 08193\\
\texttt{fanlyu@cvc.uab.cat, wenqi.zhang@autonoma.cat, joost@cvc.uab.es}\\
\textbf{Project Page}: \url{https://prefmoe.github.io/}}

\begin{document}

\maketitle

\begin{abstract}
Personalized multimodal large language models (MLLMs) aim to generate user-specific responses, but existing methods mainly rely on profile-level information and overlook diverse user preferences.
We identify group preference collapse, where multi-user personalized MLLMs become insensitive to individual preferences and drift toward dominant population-level choices due to suppressed preference signals and unreliable preference use during generation.
We propose PrefMoE, a preference-centric framework that separates stable profile information from preference-related representations.
PrefMoE decomposes preferences into shared prototypes and personalized residuals, preserves individualized residuals with imbalance-aware learning, counterfactual pseudo-user augmentation, and residual decorrelation, and routes profile and preference factors through separate LoRA adaptation paths.
Experiments across multiple MLLM backbones show that PrefMoE improves preference-sensitive personalization while substantially reducing preference collapse.
\end{abstract}

\vspace{-5bp}
\section{Introduction}
\vspace{-5bp}

\begin{figure}[!b]
    \vspace{-10bp}
    \centering
    \includegraphics[width=1\linewidth]{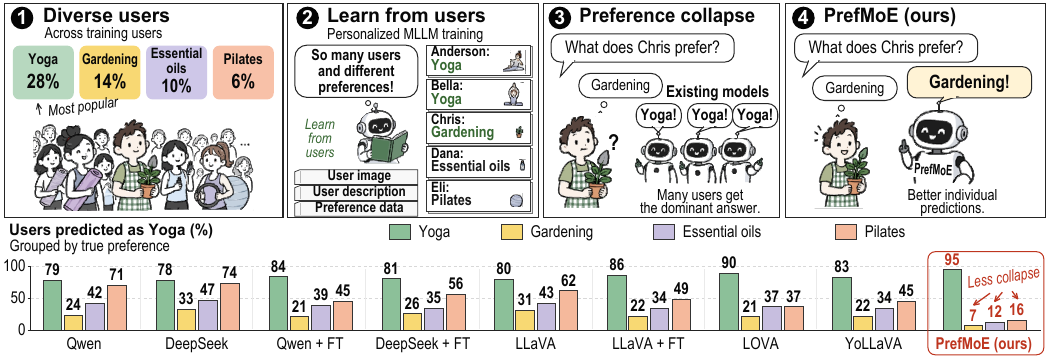}
    \vspace{-20px}
    \caption{Illustration of \emph{group preference collapse}.
        Although users have diverse preferences, existing personalized MLLMs tend to predict the dominant preference across users (``Yoga'') rather than reflecting the target user's specific preference.
        Bars indicate the percentage of users from each ground-truth preference group whose Top-1 preference is the dominant preference (``Yoga'').
    }
    \label{fig1:collapse}
\end{figure}

Multimodal Large Language Models (MLLMs)~\citep{alayrac2022flamingo,li2023blip2,dai2023instructblip,zhu2023minigpt4} have shown strong performance across a wide range of multimodal tasks, including visual question answering~\citep{Antol_2015_ICCV,fang2025guided,wang2025marten}, visual grounding~\citep{tang2026visual}, and multimodal reasoning~\citep{huang2025boosting}. 
Building on this progress, personalized MLLMs have recently attracted increasing attention~\citep{alaluf2024myvlm,cohen2022pervl,yeh2023metapersonalizing,nguyen2025yochameleon}, aiming to generate responses tailored to a target user by conditioning on user-related information~\citep{seifi2025personalizationtoolkit}.
Existing efforts mainly incorporate user-related profile signals~\citep{alaluf2024myvlm,pvit,plvm,seifi2025personalizationtoolkit}, such as portraits, demographic cues, or textual descriptions, to ground responses in user identity and attributes~\citep{pvit}. 
\textit{However, these methods primarily model who the user is, while paying much less attention to what the user prefers.} 

Preference personalization differs from profile-based personalization because preferences describe context-dependent user choices rather than relatively stable identity cues~\citep{rendle2009bpr}.
Users with similar profiles may still prefer different clothing styles, travel plans, or food options under the same visual query.
This distinction motivates dedicated mechanisms that go beyond profile modeling to preserve individual preference differences and selectively use the preference cues relevant to the current image-question pair.
Despite its practical importance, preference-level personalization remains largely underexplored.

Under joint multi-user learning, personalized MLLMs may exhibit a failure mode that we term \emph{group preference collapse}, where predictions favor population-dominant choices over individual user preferences.
As illustrated in Fig.~\ref{fig1:collapse}, this behavior persists even when user preferences are explicitly provided as textual conditions and the models are fine-tuned.
We attribute group preference collapse to two key failures.
First, individualized preference signals can be suppressed during user representation learning, as sparse and low-frequency preferences are easily absorbed by dominant population-level patterns.
Second, preference use during generation can be unreliable.
Even when preference information is provided, the model may fail to select the preference cue relevant to the current image-question pair and instead follow generic or population-dominant reasoning patterns.
Together, these failures make responses drift toward broadly plausible group-level preferences rather than fine-grained user-specific choices.

To address these failures, we propose \textit{PrefMoE}, a preference-centric personalization framework for MLLMs.
PrefMoE first learns a structured user state that separates stable profile cues from preference-related information.
It decomposes preference representations into shared population-level prototypes and personalized residuals, allowing dominant group tendencies and user-specific deviations to be modeled separately.
To prevent minority preferences from being absorbed by shared patterns, we preserve individualized residuals through imbalance-aware contrastive learning, counterfactual preference augmentation, and decorrelation.
In addition, PrefMoE introduces a factorized hierarchical MoE router to ensure that learned preferences are used during generation: preference residuals activate preference-specific LoRA experts, profile cues activate a profile-aware adaptation branch, and a query-dependent router adaptively fuses these updates for the current visual question.
In this way, PrefMoE not only preserves fine-grained preference diversity in user representations, but also turns structured user states into factorized computation paths for preference-sensitive response generation.
Our work makes three main contributions:

\begin{enumerate}[label=(\arabic*),left=0pt,itemsep=0pt]
    \vspace{-5bp}
    \item We identify group preference collapse as a critical failure mode in multi-user personalized MLLMs, where dominant population-level preferences suppress user-specific preference signals even when explicit user information is available.
    \item We propose PrefMoE, a preference-centric framework that combines residual-based preference factor learning with factorized hierarchical routing to preserve individualized preference offsets and activate profile- and preference-aware adaptation paths during generation.
    \item Experiments across multiple MLLM backbones show that PrefMoE improves preference accuracy and reduces preference collapse. On LLaVA-1.5-7B under the 0-turn setting, PrefMoE improves preference accuracy from 44.13\% to 67.33\% and reduces collapse from 34.25\% to 12.33\% compared with full fine-tuning.
\end{enumerate}

\vspace{-5bp}
\section{Related Work}
\vspace{-5bp}

\zwq{\textbf{MLLM Personalization}.
MLLM personalization associates user-specific concepts and information with visual inputs through learnable embeddings, concept tokens, prompts, and in-context conditioning~\citep{yollava,pvit}.
Recent work expands concept learning and representations~\citep{an2025unictokens}, personal memory retrieval~\citep{xu2025jarvis}, and streaming video understanding~\citep{zheng2026pearl}.
M2A also maintains and updates multimodal memories, including user preferences, during long-term interactions~\citep{feng2026m2a}.
These approaches support concept acquisition, memory access, and temporal interaction.
We focus on preference diversity under joint multi-user learning: preserving low-frequency user-specific preferences and preventing responses from drifting toward population-dominant choices.
\\
\textbf{Preference Learning}.
Preference learning models preferred outputs from comparisons, rankings, or choice-based feedback~\citep{bradley1952paired}.
Human feedback also underlies LLM alignment methods including RLHF~\citep{ouyang2022instructgpt}, DPO~\citep{rafailov2023dpo}, contrastive preference learning~\citep{hejna2024cpl}, and KTO~\citep{ethayarajh2024kto}.
Learning shared or aggregated preferences~\citep{ge2024axioms} differs from jointly modeling users with heterogeneous preferences.
Under multi-user learning, dominant preference signals can suppress individual or minority choices.
Our work investigates this failure in personalized MLLMs and combines residual preference learning with factorized routing to preserve user-specific preferences and use them during generation.
An extended discussion is provided in Appendix~\ref{app:related_work}.}
\vspace{-5bp}

\section{MLLM Personalization and Group Preference Collapse}
\vspace{-5bp}

\lyu{We study joint multi-user personalization of MLLMs, where a shared model learns from the available information of multiple users and generates personalized responses for these users at inference time.}
For each user \(u_i\), we denote the available personalization information as \(\mathcal{X}_i=\{I_i,D_i,\mathcal{P}_i\}\), where \(I_i\) and \(D_i\) denote user-related image information and textual profile description, and \(\mathcal{P}_i\) denotes the user preference set over different preference facets.
Here, preference facets refer to semantic dimensions of user preferences, such as fashion, lifestyle, travel, shopping, and entertainment.
The profile information \(I_i\) and \(D_i\) captures relatively stable identity-related cues, while \(\mathcal{P}_i\) captures user-specific tendencies across these facets and answers what the user prefers.

We take visual question answering as an example downstream task.
Given an input image \(V_n\), a question \(Q_n\), for the target user \(u_i\), the goal is to learn a personalized MLLM that is able to predict the user-specific answer \(A_{n,i}\):
\begin{equation}
\mathcal{L}^{\mathrm{vqa}}_{n,i} = \mathrm{CE}\big(\mathrm{MLLM}(V_n,Q_n;\mathcal{X}_i),A_{n,i}\big),
\end{equation}
where \(\mathrm{CE}\) denotes the answer cross-entropy prediction loss.
This formulation defines the desired user-conditioned prediction, but does not prescribe how the user condition should be implemented.

\lyu{A straightforward implementation is to encode the user's profile and preference information as tokens and concatenate them with the visual and question tokens.
However, explicitly providing user information does not guarantee that individual preferences are preserved and used appropriately.
\textit{Two challenges} arise under joint multi-user learning.
First, during representation learning, low-frequency preference signals may be overshadowed by dominant population patterns, making individual preferences difficult to distinguish.
Second, during generation, the model may rely on visual priors or common answer patterns rather than use the preference cues relevant to the current image-question pair.
These challenges can lead to responses that favor population-dominant choices over the target user's preferences.
We term this systematic tendency \emph{group preference collapse}.
Operationally, a collapse error falls outside the target user's acceptable preferences for the current image-question pair and shifts toward a more popular preference within the relevant facet.
Incorrect predictions without this population-directed shift are ordinary errors, while verified acceptable alternatives are not counted as errors.}

\vspace{-5bp}
\section{Method}
{PrefMoE addresses the two challenges discussed in the previous section through preference representation learning and query-dependent adaptation.
First, it separates shared preference structure from personalized residuals and regularizes these residuals to preserve individual preference differences.
Second, it uses hierarchical routing to combine profile- and preference-conditioned LoRA updates according to the current image-question pair, supporting the selective use of the learned user factors during generation.}

\begin{figure}[t]
    \centering
    \includegraphics[width=1\linewidth]{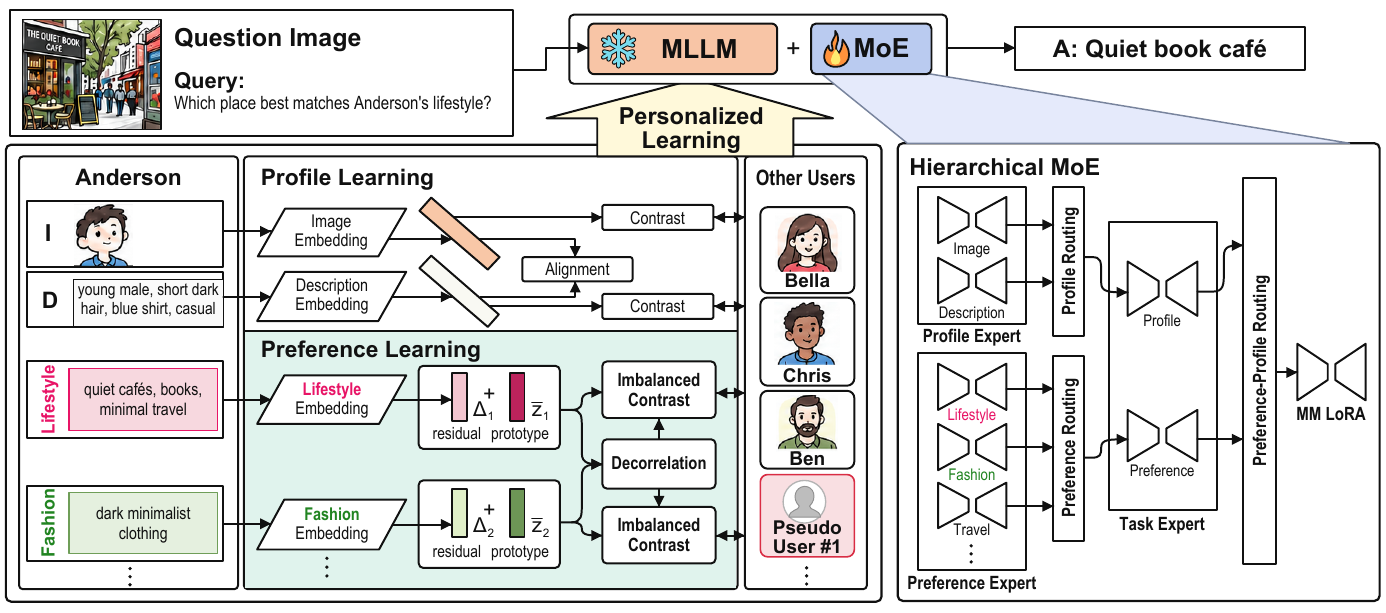}
    \vspace{-20bp}
    \caption{\textit{Overview of PrefMoE}.
PrefMoE separates profile and preference representations, models preferences with shared prototypes and personalized residuals, and regularizes the residuals with contrastive learning and decorrelation.
A hierarchical MoE router then activates profile- and preference-aware LoRA experts for query-dependent personalized reasoning.}
    \label{fig:method}
    \vspace{-15bp}
\end{figure}

\vspace{-5bp}
\subsection{Factorized User Representation and Profile Learning}
\vspace{-5bp}

Personalized responses may depend on different aspects of user information.
Profile-related questions require identity and attribute cues, whereas preference-related questions require the user's choices within the relevant semantic facets.
We therefore organize user information into separate profile and preference factors, allowing these sources to be learned and used selectively.
Given the personalization information $\mathcal{X}_i=\{I_i,D_i,\mathcal{P}_i\}$ of user $u_i$, we construct the factorized state $\{\mathbf{z}^{\mathrm{img}}_i,\mathbf{z}^{\mathrm{des}}_i,\{\mathbf{z}^{\mathrm{pref}}_{i,f}\}_{f=1}^{F}\}$.
The profile factors $\mathbf{z}^{\mathrm{img}}_i$ and $\mathbf{z}^{\mathrm{des}}_i$ capture complementary information from the user's images and textual description.
The preference factors $\mathbf{z}^{\mathrm{pref}}_{i,f}$ represent the user's preferences across $F$ semantic facets, such as lifestyle and fashion.
This factorization makes the different sources of user information separately accessible to the subsequent learning objectives and hierarchical router.

We first learn the profile factors from the visual and textual inputs:
\begin{equation}
\mathbf{z}^{\mathrm{img}}_i = g_{\mathrm{img}}(I_i),
\quad
\mathbf{z}^{\mathrm{des}}_i = g_{\mathrm{des}}(D_i),
\end{equation}
where $g_{\mathrm{img}}$ and $g_{\mathrm{des}}$ are the visual and textual profile encoders.
Because the two inputs describe the same user, we encourage cross-modal alignment while preserving distinctions between users.
The profile learning objective is
\begin{equation}
\mathcal{L}_{\mathrm{prof}}=\mathcal{L}_{\mathrm{align}}^{\mathrm{prof}}+\mathcal{L}_{\mathrm{contra}}^{\mathrm{prof}}.
\end{equation}
Here, $\mathcal{L}_{\mathrm{align}}^{\mathrm{prof}}$ aligns the visual and textual profile factors of the same user, while $\mathcal{L}_{\mathrm{contra}}^{\mathrm{prof}}$ encourages discrimination between users.
These profile factors provide identity-related information for personalization, while the preference factors are learned through the following residual-based formulation.

\vspace{-5bp}
\subsection{Preserving Individual Preference Residuals}
\vspace{-5bp}

Under joint multi-user learning, frequent preference patterns may overshadow less common user-specific signals.
To preserve individual preference differences, we decompose each facet-specific preference factor into a shared prototype and a personalized residual:
\begin{equation}
\mathbf{z}_{i,f}^{\mathrm{pref}}=\bar{\mathbf{z}}_{f}+\Delta_{i,f},\qquad f=1,\dots,F,
\end{equation}
where $\bar{\mathbf{z}}_f$ is shared across users within facet $f$, and $\Delta_{i,f}$ represents the preference offset of user $u_i$ from this shared reference.
We obtain each personalized residual $\Delta_{i,f}$ through a learnable embedding layer and optimize it during training.
For each facet $f$, the shared prototype $\bar{\mathbf{z}}_f$ is computed by averaging the corresponding user representations across training users.
This decomposition provides separate components for shared preference structure and individual deviations.
However, the decomposition alone does not ensure that the residuals retain informative preference differences.
We therefore combine imbalance-aware residual contrast, pseudo-user augmentation, and residual decorrelation to regularize preference learning.

\textbf{Imbalance-aware Residual Contrast.}
Low-frequency preferences may be underrepresented during training.
To preserve their influence on query-specific responses, we apply an imbalance-aware contrastive objective to residual-conditioned activations rather than directly to the residual embeddings.
For each image-question pair $(V_n,Q_n)$, let $\mathbf{h}_n=\mathrm{Backbone}(V_n,Q_n)$ denote the backbone hidden representation.
For user $u_i$ and facet $f$, we compute $\mathbf{a}_{i,f}^{n}=E_f(\mathbf{h}_n;\Delta_{i,f})$, where $E_f$ produces an activation conditioned on the current visual-question context and the personalized residual.

We construct contrastive supervision using facet-level preference annotations.
Let $\mathcal{B}$ denote the candidate users in the mini-batch, and let $\mathcal{N}_{i,f}\subseteq\mathcal{B}$ contain users whose annotations match those of $u_i$ on facet $f$.
For the same image-question pair, we compute the normalized similarity mass assigned to matching preferences:
\begin{equation}
p_{i,f}^{n}=\frac{\sum_{u_j\in\mathcal{N}_{i,f}}\exp \left( \mathrm{cos}(\mathbf{a}_{i,f}^{n},\mathbf{a}_{j,f}^{n})/\tau \right)}{\sum_{u_j\in\mathcal{B}}\exp \left( \mathrm{cos}(\mathbf{a}_{i,f}^{n},\mathbf{a}_{j,f}^{n})/\tau \right)}.
\end{equation}
$\mathrm{cos}(\cdot,\cdot)$ denotes cosine similarity and $\tau$ is the temperature.
Maximizing this quantity encourages similar activations for matching facet-level preferences relative to non-matching preferences.
Inspired by focal loss~\citep{lin2017focal} and class-imbalanced reweighting~\citep{cui2019class}, we define: 
\begin{equation}
\mathcal{L}_{\mathrm{res}}=-\frac{1}{BF}\sum\nolimits_{i=1}^{B}\sum\nolimits_{f=1}^{F}(1-\rho_{i,f})^{\eta}(1-p_{i,f}^{n})\log p_{i,f}^{n},
\end{equation}
where $B=|\mathcal{B}|$, $\rho_{i,f}\in[0,1]$ is the normalized frequency of the preference group associated with $u_i$ under facet $f$, and $\eta\geq0$ controls the frequency weighting.
The factor $(1-\rho_{i,f})^{\eta}$ gives relatively greater weight to less frequent preference groups, while $(1-p_{i,f}^{n})$ emphasizes cases where matching preferences receive low similarity mass.

\textbf{Counterfactual Pseudo-user Augmentation.}
Observed users provide a limited set of preference combinations within each facet.
To expand these combinations, we construct pseudo users by recombining preference items from different real users.
Specifically, for each of the fixed $F$ facets, we randomly sample multiple preference items from other users and combine them into the pseudo user's preferences for that facet.
The resulting preference profile $\tilde{\mathcal{P}}$ contains new combinations of existing items without introducing new facets or preference values.
We include these pseudo users as additional candidates in the residual contrastive objective, with preference matches determined from their recombined facet contents.
Pseudo users participate only in residual contrastive learning and require no additional answer-level labels.

\textbf{Preference Residual Decorrelation.}
The residual contrastive objective encourages preference-specific activations but does not explicitly discourage overlapping residual directions across facets.
To reduce this redundancy, we introduce soft decorrelation penalties between different facet residuals and between each residual and its corresponding shared prototype.
Let $\tilde{\Delta}_{i,f}=\Delta_{i,f}/\|\Delta_{i,f}\|_2$ and $\tilde{\mathbf{z}}_f=\bar{\mathbf{z}}_f/\|\bar{\mathbf{z}}_f\|_2$ denote the normalized residual and prototype, respectively.
We define
\begin{equation}
\mathcal{L}_{\mathrm{dec}}(u_i) =
\frac{1}{\binom{F}{2}}
\sum\nolimits_{1 \le f < g \le F}
\left(\tilde{\Delta}_{i,f}^{\top}\tilde{\Delta}_{i,g}\right)^2
+
\frac{1}{F}
\sum\nolimits_{f=1}^F
\left(\tilde{\Delta}_{i,f}^{\top}\tilde{\mathbf{z}}_f\right)^2 .
\end{equation}
The first term encourages distinct residual directions across preference facets.
The second discourages overlap with the shared prototype direction, encouraging residuals to encode complementary information.
Both terms act as soft regularizers rather than enforcing independence between preference factors.
We average $\mathcal{L}_{\mathrm{dec}}(u_i)$ over real users in the mini-batch to obtain $\mathcal{L}_{\mathrm{dec}}$.
The overall preference learning objective combines query-conditioned residual contrast with residual decorrelation:
\begin{equation}
    \mathcal{L}_{\mathrm{pref}}=\mathcal{L}_{\mathrm{res}}+\mathcal{L}_{\mathrm{dec}}.
\end{equation}

\vspace{-5bp}
\subsection{Query-Dependent Preference Utilization}
\vspace{-5bp}

Preserving individual preferences does not ensure that the model uses the cues relevant to a given image-question pair.
We therefore introduce hierarchical routing over LoRA experts to support query-dependent use of the factorized user state.
The router first combines experts within the profile and preference branches, then fuses the two branch updates.
Given an image-question pair $(V,Q)$, we obtain query features $\mathbf{q}=[\phi_v(V);\phi_q(Q)]$ for routing.
For clarity, we describe the adaptation at a single adapter insertion site, where $\mathbf{h}$ denotes the backbone hidden representation.
Each expert is implemented as a LoRA and produces an additive low-rank update to $\mathbf{h}$.
The target user's profile factors $\mathbf{z}^{\mathrm{img}}_i,\mathbf{z}^{\mathrm{des}}_i$ and preference factors $\{\mathbf{z}_{i,f}^{\mathrm{pref}}\}_{f=1}^{F}$ provide the user conditions for adaptation.

The preference branch contains $F$ facet-specific LoRA experts $\{E_f^{\mathrm{pref}}\}_{f=1}^{F}$, while the profile branch contains two experts $\{E_{\mathrm{img}}^{\mathrm{prof}},E_{\mathrm{des}}^{\mathrm{prof}}\}$ associated with the visual and textual profile factors.
The routers assign mixture weights using the query and the corresponding user factors:
\begin{equation}
\boldsymbol{\alpha}^{\mathrm{pref}}=\sigma\left(\left\{\mathbf{w}_{\mathrm{pref},f}^\top[\mathbf{q};\mathbf{z}_{i,f}^{\mathrm{pref}}]\right\}_{f=1}^{F}\right),
\quad
\boldsymbol{\alpha}^{\mathrm{prof}}=\sigma\left(\left\{\mathbf{w}_{\mathrm{prof},k}^\top[\mathbf{q};\mathbf{z}_{i,k}^{\mathrm{prof}}]\right\}_{k\in\{\mathrm{img},\mathrm{des}\}}\right),
\end{equation}
where $\sigma$ denotes Softmax over the experts within each branch, $\mathbf{z}_{i,\mathrm{img}}^{\mathrm{prof}}=\mathbf{z}^{\mathrm{img}}_i$, and $\mathbf{z}_{i,\mathrm{des}}^{\mathrm{prof}}=\mathbf{z}^{\mathrm{des}}_i$.
The scoring vectors $\mathbf{w}_{\mathrm{pref},f}$ and $\mathbf{w}_{\mathrm{prof},k}$ are expert-specific, allowing the query to affect the relative expert weights.
The preference router weights the facet-specific experts, while the profile router weights the visual and textual profile experts.
The factor-aware MoE updates are computed as
\begin{equation}
\mathbf{h}^{\mathrm{pref}}=\sum\nolimits_{f=1}^{F}\alpha_f^{\mathrm{pref}}E_f^{\mathrm{pref}}(\mathbf{h};\mathbf{z}_{i,f}^{\mathrm{pref}}),
\quad
\mathbf{h}^{\mathrm{prof}}=\sum\nolimits_{k\in\{\mathrm{img},\mathrm{des}\}}\alpha_k^{\mathrm{prof}}E_k^{\mathrm{prof}}(\mathbf{h};\mathbf{z}_{i,k}^{\mathrm{prof}}).
\end{equation}
Here, each expert returns only the adaptation update, excluding the original backbone representation.

A second router determines the relative contributions of the profile and preference branches using the query and their updates:
\begin{equation}
\boldsymbol{\gamma}=\sigma\left(\mathbf{W}_{\mathrm{br}}[\mathbf{q};\mathbf{h}^{\mathrm{prof}};\mathbf{h}^{\mathrm{pref}}]\right).
\end{equation}
Softmax is applied over the two branches, and the adapted representation is
\begin{equation}
\tilde{\mathbf{h}}=\mathbf{h}+\gamma_{\mathrm{prof}}\mathbf{h}^{\mathrm{prof}}+\gamma_{\mathrm{pref}}\mathbf{h}^{\mathrm{pref}}.
\end{equation}
This hierarchy supports query-dependent weighting of individual user factors within each branch and adaptive fusion of profile and preference information.

\textbf{Training objective and inference.}
The overall training objective combines three loss functions:
\begin{equation}
\mathcal{L}=\mathcal{L}_{\mathrm{vqa}}+\mathcal{L}_{\mathrm{prof}}+\mathcal{L}_{\mathrm{pref}}.
\end{equation}
At inference time, the model retrieves the learned profile and preference factors associated with the target user identifier.
For each image-question pair, the hierarchical routers compute the mixture weights and combine the corresponding LoRA updates to generate a personalized response.

\section{Experiment}
\vspace{-5bp}
\subsection{Experimental Setup}
\vspace{-5bp}

\zwq{\textbf{Dataset and implementation details.}
We evaluate on MMPB-Clean, a cleaned split of MMPB~\citep{kimmmpb} covering 50 human users, 61 non-human concepts, and five preference facets.
We retain the original user-concept annotations and labels for recognition and multiple-choice questions, reconstructing preference yes/no labels when needed to reduce question-format shortcuts.
All methods share the same split and are evaluated without dialogue history (0-turn) or with a ten-turn history (10-turn).
Construction details and statistics are provided in Appendix~\ref{app:dataset}.
}
\zwq{
We implement PrefMoE in PyTorch and train it on four NVIDIA A100 GPUs using AdamW~\citep{loshchilovdecoupled}.
Training configurations are detailed in Appendix~\ref{app:implementation_details}.
}
\\
\textbf{Metrics.}
We report exact-match accuracy on profile and preference questions, together with overall accuracy.
To distinguish population-directed errors from ordinary errors, we additionally measure preference collapse:
\vspace{-5px}
\begin{equation}
\label{eq:preference_collapse}
\zwq{
\mathrm{Collapse}
=
\frac{1}{|\mathcal{O}|}
\sum\nolimits_{i\in\mathcal{O}}
\hat{y}_i\,
\mathbb{I}\!\left[
\operatorname{pop}(p_i)>
\operatorname{pop}(p_i^\star)
\right].
}
\vspace{-5px}
\end{equation}
where $\mathcal{O}$ contains binary queries about items outside the target user's annotated acceptable preference set, and $\hat{y}_i=1$ indicates that the model accepts the queried item $p_i$ as preferred.
A false acceptance counts only if $p_i$ is more popular than a fixed acceptable reference $p_i^\star$ for the same user and facet, with popularity measured by annotated user counts.
Lower is better.
Reference selection, popularity computation, and answer normalization are detailed in Appendix~\ref{app:metrics}.
\\
\textbf{Compared methods.}
We compare with representative general-purpose MLLMs across different model scales, inclduing LLaVA series, DeepSeek VL2 and Qwen2.5-VL.
We also include recent personalized or multimodal baselines, including Yo'LLaVA~\citep{yollava}, LLaVA-NeXT~\citep{liu2024llavanext}, LOVA3~\citep{zhao2024lova3}, and TG-LLaVA~\citep{yan2025tgllava}.
For general-purpose MLLMs, we report no-training (NT) and full fine-tuning (FFT) results.
For personalized multimodal methods and our method, we report PEFT-based personalization results.
At inference time, NT and FFT baselines receive explicit user profile and preference descriptions as contextual prompts, while PrefMoE only receives the user identifier and retrieves the learned structured user state from memory.

\begin{table}[t]
\centering
\caption{Major comparisons with SOTAs under 0-turn and 10-turn settings.}
\vspace{-10px}
\resizebox{\linewidth}{!}{
\begin{tabular}{llcccccccc}
\toprule
\multirow{2}{*}{\textbf{Method}} 
& \multirow{2}{*}{\textbf{Type}} 
& \multicolumn{4}{c}{\textbf{0-turn}} 
& \multicolumn{4}{c}{\textbf{10-turn}} \\
\cmidrule(lr){3-6} \cmidrule(lr){7-10}
& 
& Overall$\uparrow$ & {Preference}$\uparrow$ & {Profile}$\uparrow$ & {Collapse}$\downarrow$
& Overall$\uparrow$ & {Preference}$\uparrow$ & {Profile}$\uparrow$ & {Collapse}$\downarrow$ \\
\midrule

LLaVA-1.5-7B
& \textit{NT}
& 0.3601 & 0.3387 & 0.3716 & 0.6228
& 0.3243 & 0.3067 & 0.3337 & 0.6164 \\

LLaVA-1.5-13B
& \textit{NT}
& 0.4023 & 0.3653 & 0.4222 & 0.6027 
& 0.3720 & 0.3333 & 0.3928 & 0.6119 \\

LLaVA-OV-72B 
& \textit{NT}
& 0.5100 & 0.4800 & 0.5262 & 0.5297
& 0.4620 & 0.4333 & 0.4774 & 0.5525 \\

DeepSeek-VL2-Tiny
& \textit{NT}
& 0.3683 & 0.3613 & 0.3720 & 0.7169 
& 0.3497 & 0.3440 & 0.3527 & 0.6895 \\

DeepSeek-VL2-Small
& \textit{NT}
& 0.3866 & 0.3853 & 0.3871 & 0.6895
& 0.3668 & 0.3667 & 0.3669 & 0.6712 \\

DeepSeek-VL2
& \textit{NT}
& 0.4620 & 0.4533 & 0.4667 & 0.5890
& 0.4103 & 0.4067 & 0.4122 & 0.5982 \\

Qwen2.5-VL-7B
& \textit{NT}
& 0.4401 & 0.4000 & 0.4616 & 0.6484
& 0.4007 & 0.3433 & 0.4315 & 0.6530 \\

Qwen2.5-VL-32B
& \textit{NT}
& 0.4914 & 0.4333 & 0.5226 & 0.5708
& 0.4410 & 0.4027 & 0.4616 & 0.5982 \\

Qwen2.5-VL-72B
& \textit{NT}
& 0.5301 & 0.4600 & 0.5677 & 0.5023
& 0.5016 & 0.4453 & 0.5319 & 0.5525\\

\midrule

LLaVA-1.5-7B
& \textit{FFT}
& 0.5072 & 0.4413 & 0.5427 & 0.3425 
& 0.5026 & 0.4413 & 0.5355 & 0.3470 \\

LLaVA-1.5-13B
& \textit{FFT}
& 0.5301 & 0.4600 & 0.5677 & 0.3105 & 0.5100 & 0.4400 & 0.5477 & 0.3196 \\

LLaVA-OV-72B
& \textit{FFT}
& 0.6200 & 0.5947 & 0.6337 & 0.3151
& 0.5613 & 0.5320 & 0.5771 & 0.3379 \\

DeepSeek-VL2-Tiny
& \textit{FFT}
& 0.4709 & 0.4213 & 0.4975 & 0.5845
& 0.4681 & 0.4187 & 0.4946 & 0.5936\\

DeepSeek-VL2-Small
& \textit{FFT}
& 0.4862 & 0.3840 & 0.5412 & 0.6530 
& 0.4834 & 0.3747 & 0.5419 & 0.6530 \\

DeepSeek-VL2
& \textit{FFT}
& 0.5814 & 0.5093 & 0.6201 & 0.4521 
& 0.5748 & 0.4987 & 0.6158 & 0.4566 \\

Qwen2.5-VL-7B
& \textit{FFT}
& 0.4452 & 0.3013 & 0.5226 & 0.6530 
& 0.4807 & 0.3707 & 0.5398 & 0.7854 \\

Qwen2.5-VL-32B
& \textit{FFT}
& 0.5718 & 0.5480 & 0.5849 & 0.4795 
& 0.4914 & 0.4333 & 0.5226 & 0.5114 \\

Qwen2.5-VL-72B
& \textit{FFT}
& 0.6014 & 0.5947 & 0.6057 & 0.4018 
& 0.5217 & 0.5053 & 0.5305 & 0.3927 \\

Yo'LLaVA
& \textit{PEFT}
& 0.5040 & 0.4880 & 0.5125 & 0.2075 
& 0.4840 & 0.4680 & 0.4925 & 0.2146 \\

LLaVA-NeXT-34B
& \textit{PEFT}
& 0.5599 & 0.6200 & 0.5276 & 0.2466 
& 0.5299 & 0.5900 & 0.4976 & 0.2054 \\

LOVA3
& \textit{PEFT}
& 0.5329 & 0.5680 & 0.5140 & 0.4292 
& 0.4979 & 0.5330 & 0.4790 & 0.4247 \\

TG-LLaVA
& \textit{PEFT}
& 0.5413 & 0.5800 & 0.5204 & 0.2329 
& 0.5013 & 0.5400 & 0.4804 & 0.1506 \\

\midrule

PrefMoE (LLaVA-1.5-7B)       
& \textit{PEFT}
& 0.6751 & 0.6733 & 0.6760 & 0.1233 
& 0.5986 & 0.5840 & 0.6065 & 0.1553 \\

PrefMoE (LLaVA-1.5-13B)       
& \textit{PEFT}
& 0.7012 & 0.6800 & 0.7125 & 0.1416
& 0.6601 & 0.6267 & 0.6781 & 0.1370 \\

PrefMoE (LLaVA-OV-72B) 
& \textit{PEFT}
& 0.7893 & 0.7613 & 0.8043 & 0.1142
& 0.6914 & 0.6627 & 0.7068 & \textbf{0.1187} \\

PrefMoE (DeepSeek-VL2-Tiny) 
& \textit{PEFT}
& 0.6601 & 0.6467 & 0.6674 & 0.2511 
& 0.6033 & 0.5933 & 0.6086 & 0.2740 \\

PrefMoE (DeepSeek-VL2-Small) 
& \textit{PEFT}
& 0.7305 & 0.5813 & 0.8108 & 0.2740
& 0.6382 & 0.5200 & 0.7018 & 0.2694 \\

PrefMoE (DeepSeek-VL2) 
& \textit{PEFT}
& 0.7991 & 0.7520 & \textbf{0.8244} & 0.1324 
& 0.7012 & \textbf{0.6800} & 0.7125 & 0.1826 \\

PrefMoE (Qwen2.5-VL-7B)        
& \textit{PEFT}
& 0.7613 & 0.6880 & 0.8007 & 0.1781 
& 0.6503 & 0.6253 & 0.6638 & 0.1826 \\

PrefMoE (Qwen2.5-VL-32B)        
& \textit{PEFT}
& 0.7902 & 0.7640 & 0.8043 & 0.1416
& 0.6900 & 0.6653 & 0.7032 & 0.2100 \\

PrefMoE (Qwen2.5-VL-72B)        
& \textit{PEFT}
& \textbf{0.8117} & \textbf{0.7893} & 0.8237 & \textbf{0.1096}
& \textbf{0.7301} & 0.5813 & \textbf{0.8100} & 0.1279 \\
\bottomrule
\end{tabular}
}
\label{tab:major_comparisons}
\vspace{-10bp}
\end{table}

\vspace{-5bp}
\subsection{Major Comparisons and Ablation Studies}
\vspace{-5bp}

\textbf{Comparisons on MMPB-Clean}. {Table~\ref{tab:major_comparisons}} reports the main results under 0-turn and 10-turn settings.
Raw MLLMs exhibit high preference collapse, showing that general multimodal capability alone is insufficient for user-specific preference modeling.
Full fine-tuning improves accuracy but still fails to reliably preserve preference boundaries.
By contrast, our method consistently improves overall and preference accuracy while reducing collapse across backbones and model scales.
On LLaVA-1.5-7B, it improves 0-turn overall accuracy from 0.5072 to 0.6751 and reduces collapse from 0.3425 to 0.1233 over FFT.
The gains further extend to stronger backbones, with different PrefMoE variants achieving the best 0-turn overall accuracy, the best 10-turn preference accuracy, and the lowest collapse rates.
These results suggest that preference-sensitive personalization benefits from factorized adaptation rather than relying solely on scaling or standard fine-tuning.
\\
\textbf{Ablation studies}. {Table~\ref{tab:ablation_study}} shows that each component contributes to our method.
The basic user embedding still suffers from severe collapse, indicating that storing user information alone is insufficient.
Profile factor learning brings the largest early improvement, while imbalance-aware residual preservation further improves preference accuracy by protecting low-density preference offsets.
Counterfactual augmentation and preference decorrelation mainly reduce collapse, suggesting that they help break group-level shortcuts and separate preference facets.
The hierarchical MoE router achieves the best results, confirming the need to couple factorized user states with factorized computation paths.

\begin{figure}[h]
    \vspace{-10px}
    \centering
    \includegraphics[width=\linewidth]{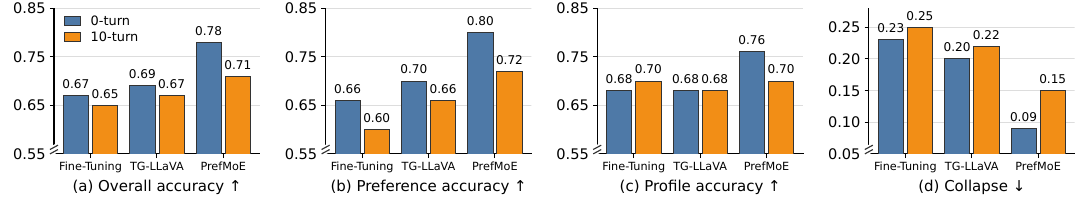}
    \vspace{-20px}
    \caption{Cross-domain results on food and fashion using LLaVA-1.5-7B.}
    \label{fig:cross_domain_validation}
    \vspace{-10px}
\end{figure}

\textbf{Validation on additional domains}.  Beyond MMPB-Clean, we further construct a validation set using images from Food-101~\citep{bossard2014food101} and Fashion Product Images (Small)~\citep{aggarwal2019fashion}.
It contains 20 users, 10 food and fashion preference categories, 500 training questions, and 100 test questions, equally divided between profile and preference questions.
The personalized questions and labels are constructed for this evaluation and independently verified by human annotators for consistency with the images and user preferences.
Fig.~\ref{fig:cross_domain_validation} shows that PrefMoE achieves the highest preference accuracy and lowest collapse among the compared methods in both dialogue settings.
These results provide complementary evidence beyond MMPB-Clean, although the small, constructed set limits broader real-world conclusions.
\begin{table}[t]
\centering
\caption{Component-wise ablation study of the proposed method. 
E, P, I, C, D, and M denote the basic user embedding module, profile factor learning, imbalance-aware residual preservation, counterfactual user augmentation, preference decorrelation, and hierarchical MoE router, respectively.}
\vspace{-10px}
\resizebox{\linewidth}{!}{
\begin{tabular}{cccccc|cccc|cccc}
\toprule
\multicolumn{6}{c|}{} 
& \multicolumn{4}{c|}{\textbf{0-turn}} 
& \multicolumn{4}{c}{\textbf{10-turn}} \\
\textbf{E} & \textbf{P} & \textbf{I} & \textbf{C} & \textbf{D} & \textbf{M} & {Overall}$\uparrow$
& {Preference}$\uparrow$
& {Profile}$\uparrow$
& {Collapse}$\downarrow$
& {Overall}$\uparrow$
& {Preference}$\uparrow$
& {Profile}$\uparrow$
& {Collapse}$\downarrow$ \\
\midrule
\cmark &  &  & & & & 0.5562 & 0.4120 & 0.6337 & 0.6027 & 0.4928 & 0.3584 & 0.5651 & 0.6347 \\
\cmark & \cmark & & &  & & 0.6000 & 0.5700 & 0.6159 & 0.3333 & 0.5266 & 0.4902 & 0.5461 & 0.3653 \\
\cmark & \cmark & \cmark & & & & 0.6279 & 0.6133 & 0.6358 & 0.3288 & 0.5506 & 0.5275 & 0.5631 & 0.3607 \\
\cmark & \cmark & \cmark & \cmark & & & 0.6303 & 0.6253 & 0.6330 & 0.2283 & 0.5571 & 0.5393 & 0.5667 & 0.2603 \\
\cmark & \cmark & \cmark & \cmark & \cmark & & 0.6382 & 0.6467 & 0.6337 & 0.1457 & 0.5688 & 0.5622 & 0.5723 & 0.1781 \\
\cmark & \cmark & \cmark & \cmark & \cmark & \cmark & \textbf{0.6751}& \textbf{0.6733}& \textbf{0.6760}& \textbf{0.1233} & \textbf{0.5986} & \textbf{0.5840} & \textbf{0.6065} & \textbf{0.1553} \\
\bottomrule
\end{tabular}
}
\label{tab:ablation_study}
\vspace{-10bp}
\end{table}

\subsection{More Analysis}
\vspace{-5bp}

\begin{figure}[h]
    \centering
    \vspace{-5bp}
    \begin{minipage}{0.3\linewidth}
        \centering
        \includegraphics[width=\linewidth]{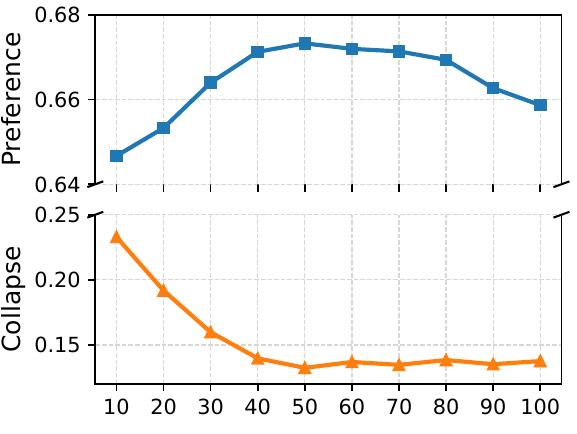}
        \vspace{-20bp}
        \caption{No. of pseudo users.}
        \label{fig:nb_user}
    \end{minipage}
    \begin{minipage}{0.3\linewidth}
        \centering
        \includegraphics[width=\linewidth]{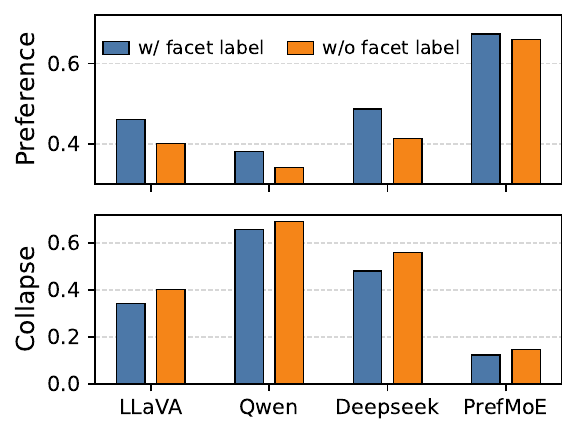}
        \vspace{-20bp}
        \caption{W/o facet labels.}
        \label{fig:no_label}
    \end{minipage}
    \begin{minipage}{0.36\linewidth}
        \centering
        \includegraphics[width=\linewidth]{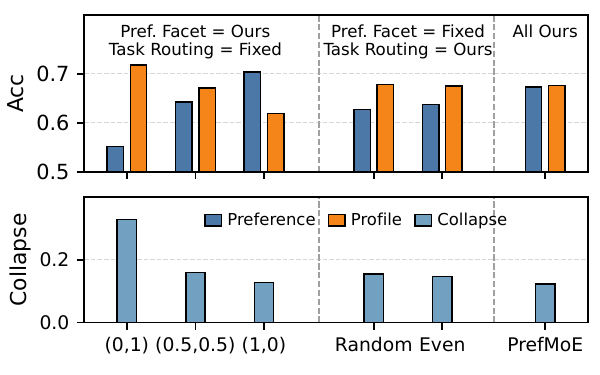}
        \vspace{-20bp}
        \caption{Routing analysis.}
        \label{fig:routing}
    \end{minipage}
    \vspace{-5bp}
\end{figure}

\textbf{Number of pseudo users}.
Fig.~\ref{fig:nb_user} shows that performance improves as the number of pseudo users increases from 10 to 50, while both Preference and Collapse become saturated afterward. This suggests that 50 pseudo users are sufficient to capture representative personalized patterns in our setting, and we therefore use this value by default.
\\
\textbf{w/o vs. w/ facet labels}.
MMPB annotations organize preferences into five facets: entertainment, travel, lifestyle, shopping, and fashion.
To test whether our method relies on explicit facet-name shortcuts, we remove these facet labels and merge all preference descriptions into a single unstructured text paragraph.
In this setting, facet structure is not directly provided and must be inferred from the preference content, mainly through the decorrelation-based factor learning.
As shown in Fig.~\ref{fig:no_label}, our method remains robust without facet labels and still outperforms FFT baselines across different backbones.
This suggests that the gains do not mainly come from surface-level facet names, but from learning structured target-specific preference relations from the underlying descriptions.

\begin{figure}[t]
    \centering
    \begin{minipage}{0.28\linewidth}
        \centering
        \includegraphics[width=\linewidth]{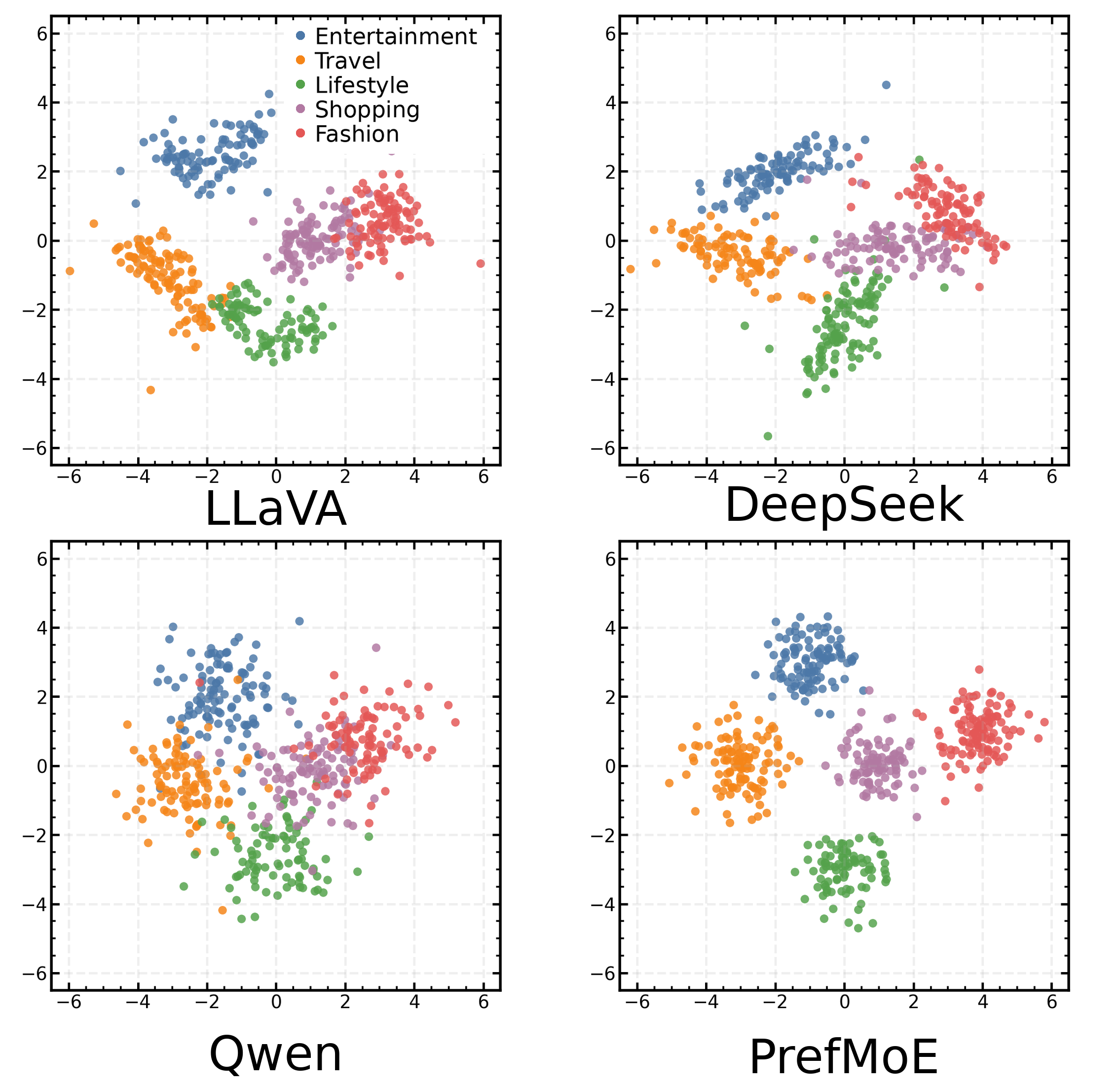}
        \vspace{-20bp}
        \caption{t-SNE of 5 different preference facets.}
        \label{fig:tsne_different_factors}
    \end{minipage}
    \hfill
    \begin{minipage}{0.69\linewidth}
        \centering
        \includegraphics[width=\linewidth]{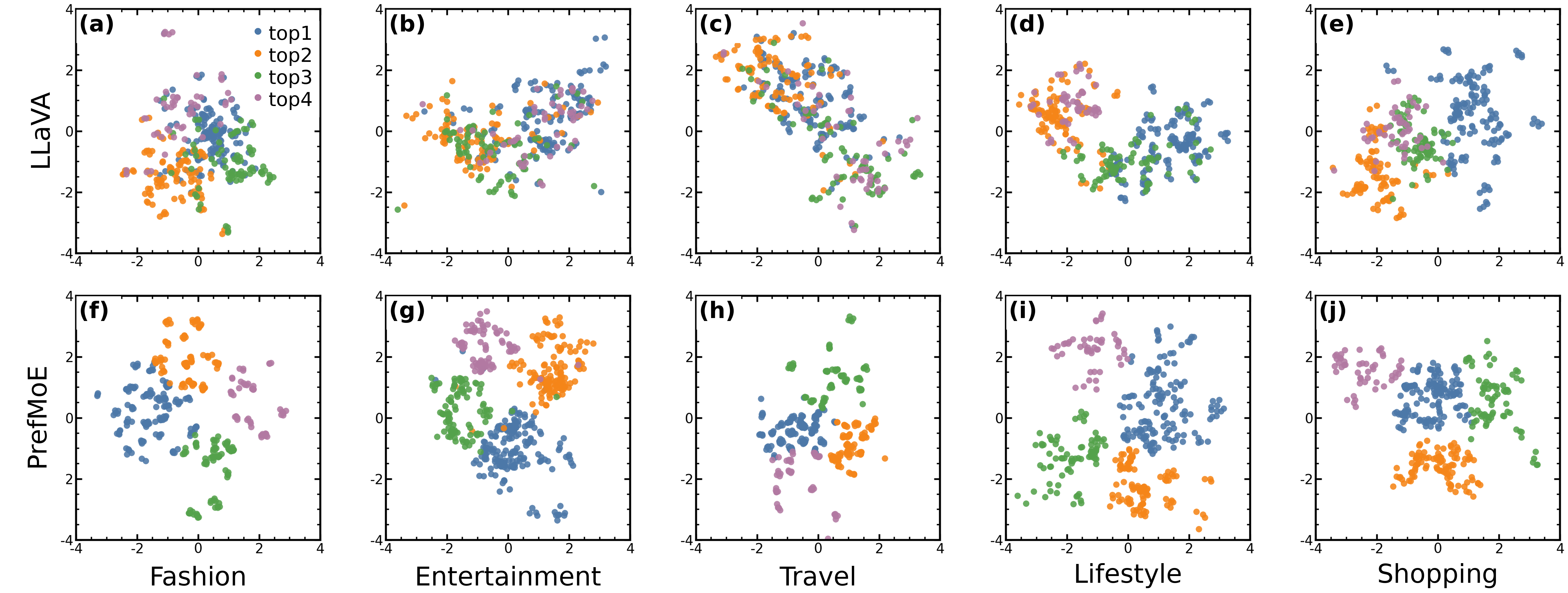}
        \vspace{-20bp}
        \caption{t-SNE visualization of top-1 to top-4 preference groups within five preference facets.}
        \label{fig:tsne_top_populations}
    \end{minipage}
    \vspace{-15bp}
\end{figure}

\begin{figure}[t]
    \centering
    \includegraphics[width=1\linewidth]{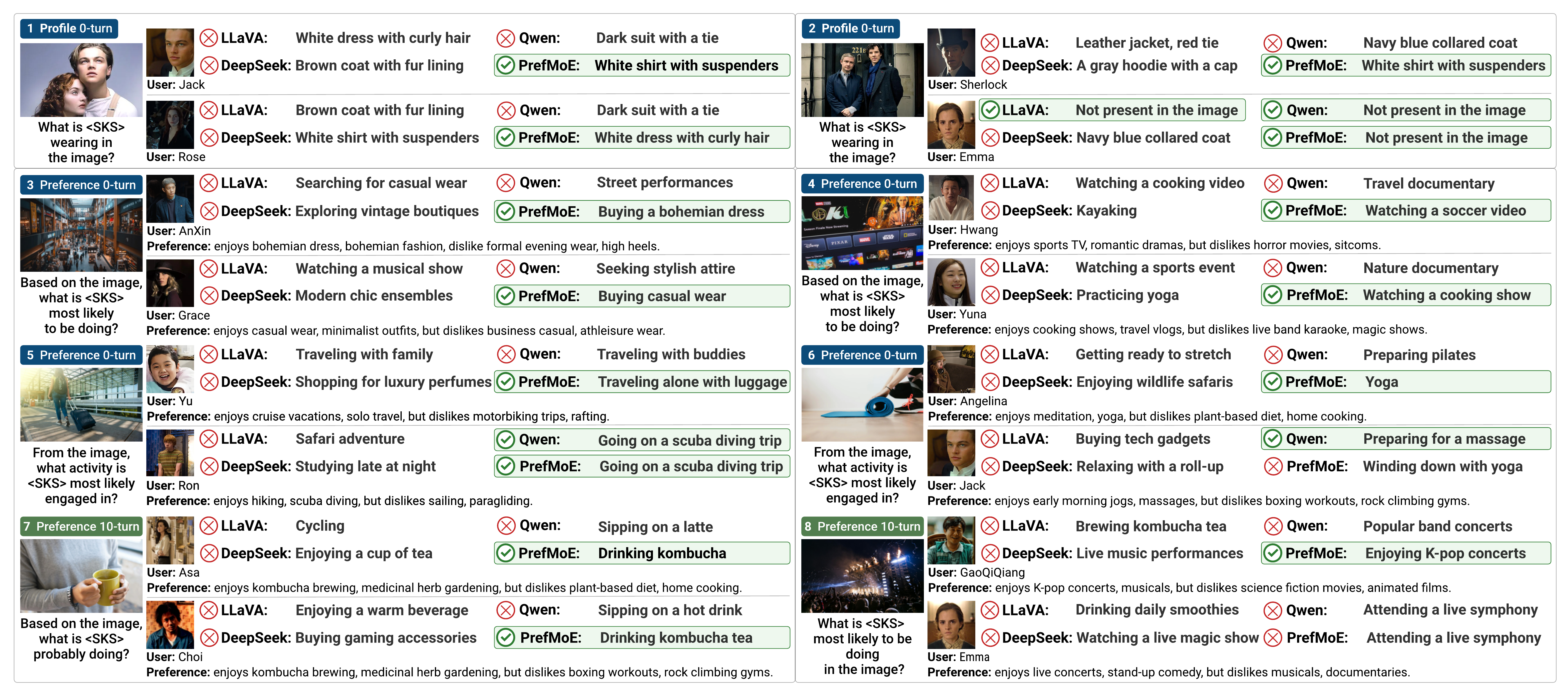}
    \vspace{-20bp}
    \caption{Qualitative examples. \texttt{<SKS>} denotes the target personalized identity or concept. Green boxes indicate correct predictions, while red marks indicate incorrect predictions. }
    \label{fig:case_study}
    \vspace{-15bp}
\end{figure}

\textbf{Adaptive routing vs. fixed routing.}
Fig.~\ref{fig:routing} analyzes the routing behavior.
We first fix the preference--profile branch weights while keeping our preference facet router unchanged.
The results show a clear trade-off: profile-only routing achieves strong profile accuracy but weak preference accuracy and severe collapse, whereas preference-only routing improves preference accuracy and reduces collapse but sacrifices profile accuracy and overall performance.
Although balanced fixed routing mitigates this trade-off, it still underperforms our adaptive branch router.
We also replace the adaptive preference facet router with random or uniform facet selection while keeping the adaptive branch router.
Both variants degrade performance, showing that query-dependent facet selection is necessary.
Overall, PrefMoE achieves the best overall accuracy and the lowest collapse in the 0-turn setting, suggesting that effective personalized reasoning requires both adaptive profile--preference fusion and adaptive preference facet routing.
\\
\textbf{t-SNE visualizations.}
We use t-SNE to qualitatively examine whether the learned preference representations preserve structured preference boundaries at both facet and population levels.
Fig.~\ref{fig:tsne_different_factors} visualizes representations from the five semantic facets, including entertainment, travel, lifestyle, shopping, and fashion.
Compared with standard fine-tuning, our method produces more separated facet-level distributions, suggesting that preference decorrelation helps different facets capture complementary user-specific offsets rather than collapsing into redundant directions.
Fig.~\ref{fig:tsne_top_populations} further examines the top-1 to top-4 preference populations within each facet.
Standard fine-tuning tends to mix these populations into shared regions, indicating that frequent and semantically related preferences are not well distinguished.
In contrast, our method preserves clearer local population structures within the same broad facet.
These results suggest that our method mitigates preference collapse not only by separating different preference facets, but also by maintaining fine-grained boundaries among preference populations inside each facet.
\\
\textbf{Qualitative Case Study}.
Fig.~\ref{fig:case_study} presents representative profile and preference cases under 0-turn and 10-turn settings.
For profile questions, baseline MLLMs often rely on salient visual cues and produce target-agnostic descriptions, while PrefMoE better conditions on the queried user.
For preference questions, PrefMoE more often predicts answers consistent with individual preferences, such as bohemian fashion, solo travel, kombucha, and K-pop concerts.
The failure cases also reveal remaining challenges.
In Case 6, PrefMoE identifies a relevant preference but fails to precisely ground the visual activity, while Case 8 shows confusion among fine-grained entertainment categories.
These results highlight both the effectiveness and limitations of preference-grounded personalized reasoning.

\vspace{-5bp}
\section{Conclusion}
\vspace{-5bp}
In this paper, we identified group preference collapse as a critical failure mode in multi-user personalized MLLMs, where responses drift toward dominant population-level preferences rather than reflecting individualized user preferences.
We analyzed this problem from both representation and reasoning perspectives, showing that sparse preference signals can be suppressed in user representations and unreliably used during generation.
To address this issue, we proposed PrefMoE, a preference-centric personalization framework that separates stable profile information from preference representations and explicitly preserves individualized preference residuals. A hierarchical routing mechanism further directs profile and preference factors into separate LoRA adaptation paths for query-dependent personalized reasoning. Experiments across multiple MLLM backbones, dialogue settings, and additional visual domains show that PrefMoE consistently improves preference-sensitive personalization and reduces preference collapse.

Despite these improvements, PrefMoE still relies on explicit and well-specified preference descriptions. Personalized reasoning may remain difficult when preferences are underspecified, weakly grounded in the visual scene, or not clearly relevant to the query. The additional validation set is also small and controlled, with a bounded preference vocabulary. Although PrefMoE maintains higher preference accuracy under the evaluated sequential-update setting, our current results do not establish improved resistance to forgetting, and longer-term continual adaptation remains an open challenge.

\bibliographystyle{iclr2027_conference}
\bibliography{ref}

@inproceedings{alaluf2024myvlm,
  title={{MyVLM}: Personalizing {VLMs} for User-Specific Queries},
  author={Yuval Alaluf and Elad Richardson and Sergey Tulyakov and Kfir Aberman and Daniel Cohen-Or},
  booktitle={ECCV},
  pages={73--91},
  year={2024}
}

@article{tang2026visual,
  title={Visual Position Prompt for {MLLM} Based Visual Grounding},
  author={Wei Tang and Yanpeng Sun and Qinying Gu and Zechao Li},
  journal={IEEE Transactions on Multimedia},
  year={2026},
  doi={10.1109/tmm.2026.3654372},
  volume={28},
  pages={3739--3754}
}

@inproceedings{yollava,
  title={{Yo'LLaVA}: Your Personalized Language and Vision Assistant},
  author={Thao Nguyen and Haotian Liu and Yuheng Li and Mu Cai and Utkarsh Ojha and Yong Jae Lee},
  booktitle={NeurIPS},
  volume={37},
  year={2024},
  pages={40913--40951},
  doi={10.52202/079017-1294}
}

@inproceedings{pvit,
  title={Personalized Visual Instruction Tuning},
  author={Renjie Pi and Jianshu Zhang and Tianyang Han and Jipeng Zhang and Rui Pan and Tong Zhang},
  booktitle={ICLR},
  year={2025},
  pages={43247--43266}
}

@inproceedings{plvm,
  title={{PLVM}: A Tuning-Free Approach for Personalized Large Vision-Language Model},
  author={Chau Pham and Hoang Phan and David Doermann and Yunjie Tian},
  booktitle={CVPR Workshops},
  pages={3671--3680},
  year={2025}
}

@article{mcllava,
  title={{MC-LLaVA}: Multi-Concept Personalized Vision-Language Model},
  author={An, Ruichuan and Yang, Sihan and Zhang, Renrui and Lu, Ming and Jiang, Tianyi and Zeng, Kai and Luo, Yulin and Cao, Jiajun and Liang, Hao and Chen, Ying and She, Qi and Zhang, Shanghang and Zhang, Wentao},
  journal={arXiv preprint arXiv:2411.11706},
  year={2024}
}

@article{onlinepvlm,
  title={{Online-PVLM}: Advancing Personalized {VLMs} with Online Concept Learning},
  author={Huiyu Bai and Runze Wang and Zhuoyun Du and Yiyang Zhao and Fengji Zhang and Haoyu Chen and Xiaoyong Zhu and Bo Zheng and Xuejiao Zhao},
  journal={arXiv preprint arXiv:2511.20056},
  year={2025}
}

@inproceedings{li2016persona,
  title={A Persona-Based Neural Conversation Model},
  author={Jiwei Li and Michel Galley and Chris Brockett and Georgios P. Spithourakis and Jianfeng Gao and Bill Dolan},
  booktitle={ACL},
  pages={994--1003},
  year={2016},
  doi={10.18653/v1/P16-1094}
}

@inproceedings{zhang2018personachat,
  title={Personalizing Dialogue Agents: {I} Have a Dog, Do You Have Pets Too?},
  author={Saizheng Zhang and Emily Dinan and Jack Urbanek and Arthur Szlam and Douwe Kiela and Jason Weston},
  booktitle={ACL},
  pages={2204--2213},
  year={2018},
  doi={10.18653/v1/P18-1205}
}

@inproceedings{chi2017speaker,
  title={Speaker Role Contextual Modeling for Language Understanding and Dialogue Policy Learning},
  author={Ta-Chung Chi and Po-Chun Chen and Shang-Yu Su and Yun-Nung Chen},
  booktitle={IJCNLP},
  pages={163--168},
  year={2017}
}

@article{bradley1952paired,
  title={Rank Analysis of Incomplete Block Designs: I. The Method of Paired Comparisons},
  author={Ralph Allan Bradley and Milton E. Terry},
  journal={Biometrika},
  volume={39},
  number={3--4},
  pages={324--345},
  year={1952},
  doi={10.1093/biomet/39.3-4.324}
}

@inproceedings{christiano2017deep,
  title={Deep Reinforcement Learning from Human Preferences},
  author={Paul F. Christiano and Jan Leike and Tom B. Brown and Miljan Martic and Shane Legg and Dario Amodei},
  booktitle={NeurIPS},
  volume={30},
  year={2017}
}

@article{ziegler2019rlhf,
  title={Fine-Tuning Language Models from Human Preferences},
  author={Daniel M. Ziegler and Nisan Stiennon and Jeffrey Wu and Tom B. Brown and Alec Radford and Dario Amodei and Paul Christiano and Geoffrey Irving},
  journal={arXiv preprint arXiv:1909.08593},
  year={2019}
}

@inproceedings{ouyang2022instructgpt,
  title={Training Language Models to Follow Instructions with Human Feedback},
  author={Long Ouyang and Jeffrey Wu and Xu Jiang and Diogo Almeida and Carroll Wainwright and Pamela Mishkin and Chong Zhang and Sandhini Agarwal and Katarina Slama and Alex Ray and John Schulman and Jacob Hilton and Fraser Kelton and Luke Miller and Maddie Simens and Amanda Askell and Peter Welinder and Paul F. Christiano and Jan Leike and Ryan Lowe},
  booktitle={NeurIPS},
  volume={35},
  pages={27730--27744},
  year={2022},
  doi={10.52202/068431-2011}
}

@article{bai2022rlhf,
  title={Training a Helpful and Harmless Assistant with Reinforcement Learning from Human Feedback},
  author={Yuntao Bai and Andy Jones and Kamal Ndousse and Amanda Askell and Anna Chen and Nova DasSarma and Dawn Drain and Stanislav Fort and Deep Ganguli and Tom Henighan and Nicholas Joseph and Saurav Kadavath and Jackson Kernion and Tom Conerly and Sheer El-Showk and Nelson Elhage and Zac Hatfield-Dodds and Danny Hernandez and Tristan Hume and Scott Johnston and Shauna Kravec and Liane Lovitt and Neel Nanda and Catherine Olsson and Dario Amodei and Tom Brown and Jack Clark and Sam McCandlish and Chris Olah and Ben Mann and Jared Kaplan},
  journal={arXiv preprint arXiv:2204.05862},
  year={2022}
}

@inproceedings{rafailov2023dpo,
  title={Direct Preference Optimization: Your Language Model is Secretly a Reward Model},
  author={Rafael Rafailov and Archit Sharma and Eric Mitchell and Stefano Ermon and Christopher D. Manning and Chelsea Finn},
  booktitle={NeurIPS},
  volume={36},
  year={2023},
  pages={53728--53741},
  doi={10.52202/075280-2338}
}

@inproceedings{hejna2024cpl,
  title={Contrastive Preference Learning: Learning from Human Feedback without Reinforcement Learning},
  author={Joey Hejna and Rafael Rafailov and Harshit Sikchi and Chelsea Finn and Scott Niekum and W. Bradley Knox and Dorsa Sadigh},
  booktitle={ICLR},
  year={2024},
  pages={18770--18798}
}

@inproceedings{ethayarajh2024kto,
  title={Model Alignment as Prospect Theoretic Optimization},
  author={Kawin Ethayarajh and Winnie Xu and Niklas Muennighoff and Dan Jurafsky and Douwe Kiela},
  booktitle={ICML},
  volume={235},
  pages={12634--12651},
  year={2024}
}

@inproceedings{ge2024axioms,
  title={Axioms for {AI} Alignment from Human Feedback},
  author={Luise Ge and Daniel Halpern and Evi Micha and Ariel D. Procaccia and Itai Shapira and Yevgeniy Vorobeychik and Junlin Wu},
  booktitle={NeurIPS},
  volume={37},
  year={2024},
  pages={80439--80465},
  doi={10.52202/079017-2557}
}

@misc{liu2024llavanext,
  title={{LLaVA-NeXT}: Improved Reasoning, {OCR}, and World Knowledge},
  author={Haotian Liu and Chunyuan Li and Yuheng Li and Bo Li and Yuanhan Zhang and Sheng Shen and Yong Jae Lee},
  howpublished={Technical blog},
  year={2024},
  url={https://llava-vl.github.io/blog/2024-01-30-llava-next/}
}

@inproceedings{zhao2024lova3,
  title={{LOVA3}: Learning to Visual Question Answering, Asking and Assessment},
  author={Henry Hengyuan Zhao and Pan Zhou and Difei Gao and Zechen Bai and Mike Zheng Shou},
  booktitle={NeurIPS},
  volume={37},
  year={2024},
  pages={115146--115175},
  doi={10.52202/079017-3656}
}

@article{yan2025tgllava,
  title={{TG-LLaVA}: Text Guided {LLaVA} via Learnable Latent Embeddings},
  author={Dawei Yan and Pengcheng Li and Yang Li and Hao Chen and Qingguo Chen and Weihua Luo and Wei Dong and Qingsen Yan and Haokui Zhang and Chunhua Shen},
  journal={Proceedings of the AAAI Conference on Artificial Intelligence},
  volume={39},
  number={9},
  pages={9076--9084},
  year={2025}
}

@inproceedings{kimmmpb,
  title={{MMPB}: It's Time for Multi-Modal Personalization},
  author={Jaeik Kim and Woojin Kim and Woohyeon Park and Jaeyoung Do},
  booktitle={NeurIPS Datasets and Benchmarks Track},
  year={2025},
  doi={10.52202/085713-0872},
  volume={38},
  pages={29306--29358}
}

@inproceedings{alayrac2022flamingo,
  title={Flamingo: A Visual Language Model for Few-Shot Learning},
  author={Jean-Baptiste Alayrac and Jeff Donahue and Pauline Luc and Antoine Miech and Iain Barr and Yana Hasson and Karel Lenc and Arthur Mensch and Katie Millican and Malcolm Reynolds and Roman Ring and Eliza Rutherford and Serkan Cabi and Tengda Han and Zhitao Gong and Sina Samangooei and Marianne Monteiro and Jacob Menick and Sebastian Borgeaud and Andrew Brock and Aida Nematzadeh and Sahand Sharifzadeh and Mikolaj Binkowski and Ricardo Barreira and Oriol Vinyals and Andrew Zisserman and Karen Simonyan},
  booktitle={NeurIPS},
  volume={35},
  pages={23716--23736},
  year={2022},
  doi={10.52202/068431-1723}
}

@inproceedings{li2023blip2,
  title={{BLIP-2}: Bootstrapping Language-Image Pre-training with Frozen Image Encoders and Large Language Models},
  author={Junnan Li and Dongxu Li and Silvio Savarese and Steven C. H. Hoi},
  booktitle={ICML},
  pages={19730--19742},
  year={2023},
  volume={202}
}

@inproceedings{dai2023instructblip,
  title={{InstructBLIP}: Towards General-Purpose Vision-Language Models with Instruction Tuning},
  author={Wenliang Dai and Junnan Li and Dongxu Li and Anthony Meng Huat Tiong and Junqi Zhao and Weisheng Wang and Boyang Li and Pascale Fung and Steven C. H. Hoi},
  booktitle={NeurIPS},
  volume={36},
  year={2023},
  pages={49250--49267},
  doi={10.52202/075280-2142}
}

@inproceedings{zhu2023minigpt4,
  title={{MiniGPT-4}: Enhancing Vision-Language Understanding with Advanced Large Language Models},
  author={Deyao Zhu and Jun Chen and Xiaoqian Shen and Xiang Li and Mohamed Elhoseiny},
  booktitle={ICLR},
  year={2024},
  pages={18378--18394}
}

@inproceedings{cohen2022pervl,
  title={``This Is My Unicorn, {Fluffy}'': Personalizing Frozen Vision-Language Representations},
  author={Niv Cohen and Rinon Gal and Eli A. Meirom and Gal Chechik and Yuval Atzmon},
  booktitle={ECCV},
  pages={558--577},
  year={2022},
  doi={10.1007/978-3-031-20044-1_32}
}

@inproceedings{yeh2023metapersonalizing,
  title={Meta-Personalizing Vision-Language Models to Find Named Instances in Video},
  author={Chun-Hsiao Yeh and Bryan Russell and Josef Sivic and Fabian Caba Heilbron and Simon Jenni},
  booktitle={CVPR},
  pages={19123--19132},
  year={2023}
}

@inproceedings{nguyen2025yochameleon,
  title={{Yo'Chameleon}: Personalized Vision and Language Generation},
  author={Thao Nguyen and Krishna Kumar Singh and Jing Shi and Trung Bui and Yong Jae Lee and Yuheng Li},
  booktitle={CVPR},
  pages={14438--14448},
  year={2025}
}

@article{seifi2025personalizationtoolkit,
  title={Personalization Toolkit: Training Free Personalization of Large Vision Language Models},
  author={Soroush Seifi and Vaggelis Dorovatas and Matteo Cassinelli and Fabien Despinoy and Daniel Olmeda Reino and Rahaf Aljundi},
  journal={Transactions on Machine Learning Research},
  year={2026},
  issn={2835-8856},
  url={https://openreview.net/forum?id=5mbn3B0O29}
}

@inproceedings{rendle2009bpr,
  title={{BPR}: Bayesian Personalized Ranking from Implicit Feedback},
  author={Steffen Rendle and Christoph Freudenthaler and Zeno Gantner and Lars Schmidt-Thieme},
  booktitle={UAI},
  pages={452--461},
  year={2009}
}

@inproceedings{lin2017focal,
  title={Focal loss for dense object detection},
  author={Lin, Tsung-Yi and Goyal, Priya and Girshick, Ross and He, Kaiming and Doll{\'a}r, Piotr},
  booktitle={ICCV},
  pages={2980--2988},
  year={2017}
}

@inproceedings{cui2019class,
  title={Class-balanced loss based on effective number of samples},
  author={Cui, Yin and Jia, Menglin and Lin, Tsung-Yi and Song, Yang and Belongie, Serge},
  booktitle={CVPR},
  pages={9268--9277},
  year={2019}
}

@inproceedings{loshchilovdecoupled,
  title={Decoupled Weight Decay Regularization},
  author={Loshchilov, Ilya and Hutter, Frank},
  booktitle={ICLR},
  year={2019},
  url={https://openreview.net/forum?id=Bkg6RiCqY7}
}

@inproceedings{Antol_2015_ICCV,
  author={Antol, Stanislaw and Agrawal, Aishwarya and Lu, Jiasen and Mitchell, Margaret and Batra, Dhruv and Zitnick, C. Lawrence and Parikh, Devi},
  title={{VQA}: Visual Question Answering},
  booktitle={ICCV},
  month={December},
  year={2015},
  pages={2425--2433}
}

@inproceedings{fang2025guided,
  title={Notes-guided {MLLM} Reasoning: Enhancing {MLLM} with Knowledge and Visual Notes for Visual Question Answering},
  author={Fang, Wenlong and Wu, Qiaofeng and Chen, Jing and Xue, Yun},
  booktitle={CVPR},
  pages={19597--19607},
  year={2025}
}

@inproceedings{wang2025marten,
  title={Marten: Visual question answering with mask generation for multi-modal document understanding},
  author={Wang, Zining and Guan, Tongkun and Fu, Pei and Duan, Chen and Jiang, Qianyi and Guo, Zhentao and Guo, Shan and Luo, Junfeng and Shen, Wei and Yang, Xiaokang},
  booktitle={CVPR},
  pages={14460--14471},
  year={2025}
}

@inproceedings{huang2025boosting,
  title={Boosting {MLLM} Reasoning with Text-Debiased {Hint-GRPO}},
  author={Huang, Qihan and Dai, Weilong and Liu, Jinlong and He, Wanggui and Jiang, Hao and Song, Mingli and Chen, Jingyuan and Yao, Chang and Song, Jie},
  booktitle={ICCV},
  pages={4848--4857},
  year={2025}
}

@inproceedings{bossard2014food101,
  title={{Food-101} -- Mining Discriminative Components with Random Forests},
  author={Bossard, Lukas and Guillaumin, Matthieu and Van Gool, Luc},
  booktitle={ECCV},
  year={2014},
  url={https://data.vision.ee.ethz.ch/cvl/datasets_extra/food-101/},
  doi={10.1007/978-3-319-10599-4_29},
  pages={446--461}
}

@article{yang2025slc,
  title={Small-Large Collaboration: Training-efficient Concept Personalization for Large {VLM} using a Meta Personalized Small {VLM}},
  author={Yang, Sihan and Ji, Huitong and Lu, Shaolin and Chen, Jiayi and Xu, Binxiao and Lu, Ming and Zhang, Yuanxing and Dong, Wenhui and Zhang, Wentao},
  journal={arXiv preprint arXiv:2508.07260},
  year={2025}
}

@inproceedings{an2025unictokens,
  title={{UniCTokens}: Boosting Personalized Understanding and Generation via Unified Concept Tokens},
  author={An, Ruichuan and Yang, Sihan and Zhang, Renrui and Shen, Zijun and Lu, Ming and Dai, Gaole and Liang, Hao and Guo, Ziyu and Yan, Shilin and Luo, Yulin and Zou, Bocheng and Yang, Chaoqun and Zhang, Wentao},
  booktitle={NeurIPS},
  year={2025},
  pages={144638--144664},
  doi={10.52202/085713-4840},
  volume={38}
}

@inproceedings{an2025conceptastree,
  title={{Concept-as-Tree}: A Controllable Synthetic Data Framework Makes Stronger Personalized {VLMs}},
  author={An, Ruichuan and Zeng, Kai and Lu, Ming and Yang, Sihan and Zhang, Renrui and Ji, Huitong and Liang, Hao and Zhang, Wentao},
  year={2026},
  booktitle={ECCV},
  url={https://eccv.ecva.net/virtual/2026/poster/4499}
}

@article{xu2025jarvis,
  title={{Jarvis}: Towards Personalized {AI} Assistant via Personal {KV}-Cache Retrieval},
  author={Xu, Binxiao and Feng, Junyu and Lu, Shaolin and Luo, Yulin and Yan, Shilin and Liang, Hao and Lu, Ming and Zhang, Wentao},
  journal={arXiv preprint arXiv:2510.22765},
  year={2025}
}

@article{feng2026m2a,
  title={{M2A}: Multimodal Memory Agent with Dual-Layer Hybrid Memory for Long-Term Personalized Interactions},
  author={Feng, Junyu and Xu, Binxiao and Chen, Jiayi and Dai, Mengyu and Wu, Cenyang and Li, Haodong and Zeng, Bohan and Xie, Yunliu and Liang, Hao and Lu, Ming and Zhang, Wentao},
  journal={arXiv preprint arXiv:2602.07624},
  year={2026}
}

@article{zheng2026pearl,
  title={{PEARL}: Personalized Streaming Video Understanding Model},
  author={Zheng, Yuanhong and An, Ruichuan and Lin, Xiaopeng and Liu, Yuxing and Yang, Sihan and Zhang, Huanyu and Li, Haodong and Zhang, Qintong and Zhang, Renrui and Li, Guopeng and Zhang, Yifan and Li, Yuheng and Zhang, Wentao},
  journal={arXiv preprint arXiv:2603.20422},
  year={2026}
}

@misc{aggarwal2019fashion,
  author       = {Param Aggarwal},
  title        = {{Fashion Product Images (Small)}},
  year         = {2019},
  howpublished = {Kaggle},
  url          = {https://www.kaggle.com/datasets/paramaggarwal/fashion-product-images-small}
}

\clearpage
\appendix

{\Large\textbf{Appendix}}

\raggedbottom
\setlength{\parskip}{3pt plus 1pt minus 0.5pt}
\setlength{\textfloatsep}{8pt plus 2pt minus 2pt}
\setlength{\floatsep}{7pt plus 2pt minus 2pt}
\setlength{\intextsep}{6pt plus 2pt minus 1pt}
\captionsetup{font=small,skip=5pt}
\renewcommand{\topfraction}{0.95}
\renewcommand{\bottomfraction}{0.85}
\renewcommand{\textfraction}{0.07}
\renewcommand{\floatpagefraction}{0.75}
\setcounter{topnumber}{3}
\setcounter{bottomnumber}{3}
\setcounter{totalnumber}{5}

\section{Extended Related Work}
\label{app:related_work}

\textbf{MLLM Personalization}.
Early personalization studies in language models mainly adapted generated text to personas, speaker roles, or dialogue histories, focusing on stylistic and contextual consistency in text-only interactions~\citep{li2016persona,zhang2018personachat,chi2017speaker}.
With the rise of MLLMs, personalization has been extended to multimodal inputs, where most methods focus on associating user-specific or instance-specific concepts with visual entities.
For example, existing approaches bind personalized concepts to visual representations through learnable embeddings, special concept tokens, or soft prompts~\citep{yollava,mcllava}.
Recent methods further improve scalability and flexibility through multimodal in-context conditioning, reference-based identification, and online concept learning~\citep{pvit,plvm,onlinepvlm}.
Recent work broadens personalization in several complementary directions.
SLC combines a personalized small VLM with a larger model~\citep{yang2025slc}. UniCTokens unifies concept representations for understanding and generation~\citep{an2025unictokens}. Concept-as-Tree constructs synthetic concept-learning data~\citep{an2025conceptastree}.
Retrieval and memory approaches include Jarvis, which retrieves personal visual and textual KV-caches~\citep{xu2025jarvis}, and M2A, which maintains and updates multimodal memories, including user preferences, over long-term interactions~\citep{feng2026m2a}.
PEARL studies personalized streaming video understanding~\citep{zheng2026pearl}.
These directions address concept acquisition, memory access, and temporal interaction.
Our focus is preference diversity under joint multi-user learning: preserving low-frequency user-specific preferences and preventing predictions from drifting toward more popular choices.

\textbf{Preference Learning}.
Preference learning aims to infer preferred outputs from pairwise comparisons~\citep{bradley1952paired}, rankings, or choice-based feedback and has been widely used for recommendations~\citep{rendle2009bpr} and decision making~\citep{christiano2017deep}.
Beyond classical ranking settings, preference feedback has also been used to train agents~\citep{christiano2017deep} and generative models from human judgments~\citep{ziegler2019rlhf}.
It has become a central tool for aligning large language models, from RLHF~\citep{ouyang2022instructgpt,bai2022rlhf} to direct or RL-free optimization methods such as DPO~\citep{rafailov2023dpo}, contrastive preference learning~\citep{hejna2024cpl}, and KTO~\citep{ethayarajh2024kto}.
Most existing studies focus on exploiting preference signals to optimize model outputs toward shared or aggregated human preferences~\citep{ouyang2022instructgpt,bai2022rlhf,rafailov2023dpo,ge2024axioms}.
However, this objective differs from personalized preference modeling, where multiple users with heterogeneous preferences must be modeled jointly.
In such settings, aggregating preference signals can suppress minority or user-specific preferences and make different users receive increasingly similar responses.
Our work studies this failure mode in personalized MLLMs and explicitly models preference diversity to reduce group preference collapse.

\FloatBarrier
\Needspace{6\baselineskip}
\section{Dataset Construction}

\label{app:dataset}

\subsection{Why MMPB-Clean is Needed}
\label{app:why_mmpb_clean}

We adopt MMPB~\citep{kimmmpb} as the base benchmark because it provides a relatively systematic evaluation setting for multimodal personalization. MMPB contains 10,017 image-query pairs and 111 personalizable concepts spanning humans, animals, objects, and characters. Its human category further includes preference-grounded questions, making it suitable for studying both profile-centric recognition and preference-sensitive reasoning. However, the original benchmark is not specifically designed to evaluate user-conditioned personalization under shortcut-controlled settings.

To make the evaluation better reflect whether a model retrieves and applies the personalized information associated with the queried target, we derive a cleaned evaluation protocol from MMPB, termed \textbf{MMPB-Clean}. MMPB-Clean does not introduce external images, users, concepts, or preference annotations. Instead, it reuses the original MMPB resources and reorganizes them into a shortcut-controlled protocol. Specifically, we preserve the original personalized concepts, user profiles, preference annotations, images, and natural-language questions whenever possible, while adjusting the train/evaluation split and reconstructing preference yes/no labels only when necessary. The resulting number of QA pairs may differ from the original MMPB because additional preference negative pairs are constructed from existing MMPB users and preference annotations, rather than from newly collected data.

MMPB-Clean reduces two potential shortcut sources. For recognition questions, we reduce image-level shortcuts by grouping samples derived from the same underlying image and constructing the evaluation set mainly from name-sensitive groups, where changing the queried user or concept name can change the answer. Nearly uniform groups, where most queried targets lead to the same answer, are primarily kept for training support and are prevented from dominating evaluation. For preference yes/no questions, we reduce question-format shortcuts by reformulating the task as image-semantics to user-preference matching. Instead of relying on the surface polarity of the question, such as explicit negation, the reconstructed labels indicate whether the semantic meaning of the image group matches the queried user's corresponding preference field.
All methods are trained and evaluated using the same reconstructed split, labels, answer normalization, and evaluation protocol. Therefore, MMPB-Clean provides a fair and reproducible protocol for evaluating user-conditioned personalized memory, with greater emphasis on name-sensitive recognition and preference-boundary reasoning.

\subsection{Construction Protocol}
\label{app:construction_protocol}

MMPB-Clean is constructed from the original MMPB annotations. The reconstruction consists of three parts, corresponding to profile-centric recognition questions, preference yes/no questions, and multiple-choice questions.

For profile-centric recognition questions, we preserve the original task semantics and answer labels. We first group samples by their underlying image. For each image group, we examine whether changing the queried user or concept name can change the answer. Image groups satisfying this condition are treated as name-sensitive groups. The recognition evaluation set is mainly constructed from these name-sensitive groups, while nearly uniform groups are primarily kept for training support. We remove internally inconsistent image groups.
Across the dataset, the training and evaluation splits share
387 images. These include multi-person images used in recognition
questions with different queried identities. For example, a
training question may ask whether person A is present, while an
evaluation question asks whether person B is present in the same
image. Thus, the recognition split permits image reuse across
queries about different identities and is not strictly
image-disjoint.

For preference yes/no questions, we reconstruct the evaluation labels as image-semantics to user-preference matching. Each user profile contains preference fields over several facets, including entertainment, fashion, lifestyle, shopping, and travel. For each originally positive image group, we identify the stable preference elements shared by the associated users and treat them as the semantic meaning of the image group. A query is labeled positive if this semantic meaning appears in the queried user's corresponding preference field, and negative otherwise. We keep the original positive pairs and construct additional negative pairs by selecting users whose corresponding preference fields do not contain the image-related elements. Originally negative groups are kept unchanged.

For multiple-choice questions, we keep the original questions and labels unchanged. We only adjust the split by user or concept identity, so that each target contributes a limited number of multiple-choice questions to evaluation while the remaining samples are used for training. This preserves the original multiple-choice task format while reducing repeated target-specific patterns across training and evaluation.

\subsection{Human Validation of Reconstructed Labels}
\label{app:label_validation}

The preference yes/no reconstruction uses human-designed deterministic rules rather than an LLM or a learned classifier.
Within an image group, the retained preference elements are the intersection of the associated users' preference annotations. Unrelated user--image pairs satisfy a zero-overlap constraint in the relevant preference field.
The relatedness decision and question polarity determine the yes/no label, and question polarities are checked programmatically.
We manually validate 200 reconstructed samples covering 50 users, collecting 2,000 human judgments.
The reconstructed labels agree with these judgments in 99.40\% of cases, with an average confidence of 4.97 on a five-point scale (Table~\ref{tab:label_human_validation}).
This audit supports the reliability of the sampled labels, but it does not establish that every reconstructed label is correct.

\begin{table}[!htbp]
\centering
\small
\setlength{\tabcolsep}{4pt}
\renewcommand{\arraystretch}{1.12}
\caption{Human audit of reconstructed MMPB-Clean labels. Judgments denotes individual annotation decisions, not the number of annotators.}
\label{tab:label_human_validation}
\begin{tabular}{ccccc}
\toprule
Users & Samples & Judgments & Agreement $\uparrow$ & Confidence (1--5) \\
\midrule
50 & 200 & 2,000 & 99.40\% & 4.97 \\
\bottomrule
\end{tabular}
\end{table}

\subsection{Fairness and Reproducibility}
\label{app:fairness_reproducibility}

MMPB-Clean is used only to define the data split and evaluation labels. All methods are trained and evaluated on exactly the same training and evaluation sets. We do not apply method-specific filtering, relabeling, or sample selection. The reconstructed preference labels are independent of our model design and do not rely on factorized user representations, residual preference learning, or MoE routing. Thus, MMPB-Clean provides a shared evaluation protocol rather than an architecture-specific benchmark.
To ensure fair comparison, all methods use the same visual inputs, natural-language questions, target user or concept identifiers, candidate answer space, answer normalization procedure, and evaluation metrics. For questions inherited from the original benchmark, we keep the original wording and labels. For reconstructed preference yes/no questions, the image-semantics to user-preference matching rule is applied before model training and is shared by all methods.
To support reproducibility, we will release the reconstructed split files, image-group identifiers, preference-element mappings, answer-label files, and evaluation scripts. These files are sufficient to reproduce MMPB-Clean from the original MMPB annotations and to evaluate future methods under the same protocol.

\subsection{{Dataset Statistics}}

\begin{table}[!htbp]
\centering
\caption{Detailed statistics of MMPB-Clean. The same evaluation split is used for both 0-turn and 10-turn evaluation settings.}
\label{tab:mmpb_clean_detailed_stats}
\begin{tabular}{lrrr}
\toprule
Statistic & Train & Evaluation & Total \\
\midrule
\multicolumn{4}{l}{\textit{Dataset scale}} \\
QA pairs & 10371 & 2145 & 12516 \\
Images & 8272 & 2131 & 10016 \\
Human users & 50 & 50 & 50 \\
Non-human concepts & 61 & 61 & 61 \\
\midrule
\multicolumn{4}{l}{\textit{Preference QA by facet}} \\
All preference QA & 6750 & 750 & 7500 \\
Entertainment & 1341 & 159 & 1500 \\
Fashion & 1349 & 151 & 1500 \\
Lifestyle & 1357 & 143 & 1500 \\
Shopping & 1356 & 144 & 1500 \\
Travel & 1347 & 153 & 1500 \\
\midrule
\multicolumn{4}{l}{\textit{Recognition QA by concept type}} \\
All recognition QA & 3621 & 1395 & 5016 \\
Animal & 600 & 100 & 700 \\
Character & 318 & 482 & 800 \\
Human identity & 2250 & 250 & 2500 \\
Object & 453 & 563 & 1016 \\
\bottomrule
\end{tabular}
\end{table}

Table~\ref{tab:mmpb_clean_detailed_stats} summarizes the statistics of MMPB-Clean. The dataset contains 12,516 QA pairs in total, with 10,371 examples for training and 2,145 examples for evaluation. The evaluation split preserves all 50 human preference users and all 61 non-human recognition concepts, ensuring that both personalized preference reasoning and visual concept recognition are assessed under the same user/concept coverage as training.

Compared with the original MMPB, MMPB-Clean increases the number of QA pairs from 10,017 to 12,516. The increase comes from preference QA, which is expanded from 5,001 to 7,500 pairs by constructing additional negative user--image pairs using existing MMPB users and preference annotations. Recognition QA keeps the original 5,016 pairs, with only the train/evaluation split reorganized. Thus, the changed statistics come from protocol reconstruction rather than external data collection.

For preference-oriented QA, MMPB-Clean contains 7,500 examples evenly distributed across five preference facets: entertainment, fashion, lifestyle, shopping, and travel. Each facet contains 1,500 QA pairs in total, while the train/evaluation split remains approximately balanced across facets. For recognition-oriented QA, the dataset contains 5,016 examples covering animal, character, human identity, and object recognition. In this table, question subtypes are aggregated to provide a compact view of the overall data composition.

\FloatBarrier
\Needspace{6\baselineskip}
\section{Implementation Details}
\label{app:implementation_details}

We instantiate our framework on multiple MLLM backbones, including LLaVA-1.5, LLaVA-OneVision, DeepSeek-VL2, and Qwen2.5-VL. Unless otherwise specified, we use LLaVA-1.5-7B as the default backbone for ablation studies and detailed analysis. For all backbone variants, we keep the same training objective and personalized modules, while adapting the LoRA insertion modules and batch configuration to the corresponding model architecture and memory constraints.
We apply rank-64 LoRA to seven projection modules.
We insert these adapters into the top 8 decoder layers, with two insertion sites per selected layer: one after the self-attention output and one after the MLP output. This results in 16 personalized residual-adapter insertion sites. For other backbones, we use the same top-layer insertion strategy and adapt the exact number of insertion sites according to the depth of the language decoder.
All experiments are implemented in PyTorch. We use AdamW as the optimizer with a learning rate of \(2\times10^{-4}\) and a cosine learning-rate scheduler. The default batch size is 4 for LLaVA-1.5-7B. Training is performed with bfloat16 precision and TF32 enabled, and gradient checkpointing is used to reduce memory consumption. Unless otherwise noted, experiments are conducted on NVIDIA A100-40G GPUs.

\FloatBarrier
\Needspace{6\baselineskip}
\section{Evaluation Metrics}
\label{app:metrics}

We evaluate model performance using two metrics: answer accuracy and population-directed preference collapse.

\paragraph{Exact-match Answer Accuracy.}
We evaluate answer correctness with strict exact-match accuracy after normalizing both predictions and ground-truth answers into a canonical label space.
For multiple-choice questions, we normalize answers to \texttt{a,b,c,d}. For binary questions, we normalize answers to \texttt{yes,no}.
The evaluator first extracts an explicit option or yes/no answer from the generated response.
If no explicit option is found for a multiple-choice question, it further matches the response against the candidate option texts.
Predictions that cannot be mapped to a valid canonical label are counted as incorrect.
Given an evaluation set \(\mathcal{S}\), let \(g_i\) and \(\hat{g}_i\) denote the normalized ground-truth answer and prediction for sample \(i\), respectively.

We define the per-sample correctness indicator as
\[
\mathrm{hit}_i =
\mathbb{I}\left[
\hat{g}_i = g_i
\land
\hat{g}_i \neq \emptyset
\right].
\]
The exact-match accuracy on \(\mathcal{S}\) is then computed by instance-level micro-averaging:
\[
\mathrm{Acc}(\mathcal{S}) =
\frac{1}{|\mathcal{S}|}
\sum_{i\in\mathcal{S}}\mathrm{hit}_i.
\]
We report accuracy on three evaluation sets:
\[
\mathrm{Overall}=\mathrm{Acc}(\mathcal{S}_{\mathrm{all}}),\qquad
\mathrm{Preference}=\mathrm{Acc}(\mathcal{S}_{\mathrm{pref}}),\qquad
\mathrm{Profile}=\mathrm{Acc}(\mathcal{S}_{\mathrm{prof}}).
\]
Here, \(\mathcal{S}_{\mathrm{all}}\) is the full evaluation set,
\(\mathcal{S}_{\mathrm{pref}}\) contains preference-related questions,
and \(\mathcal{S}_{\mathrm{prof}}\) contains profile-recognition questions that require user-specific profile information.
Since Overall is micro-averaged over all samples, it is weighted by the subset sizes and is not necessarily the arithmetic mean of Preference and Profile.

\paragraph{Preference Collapse.}
Accuracy alone does not reveal whether a model preserves user-specific preference boundaries.
A model may answer many samples correctly while still over-expanding a user's preference range by attributing more popular, non-preferred items to that user.
We therefore report \textbf{Collapse}, which measures erroneous preference attributions outside the target user's acceptable preference set that shift toward more popular preferences.

For each binary preference sample \(i\), let \(u_i\) denote the target user and \(S_i\) denote the true preference user set of the queried item, i.e., the users for whom the item should be considered preference-matching.
Since preference questions may use different surface formulations, we first normalize yes/no answers into a unified preference-semantic space.
Let \(y_i^{*}\in\{0,1\}\) be the normalized ground-truth label and \(\hat{y}_i\in\{0,1\}\) be the normalized model prediction, where \(1\) means preferred and \(0\) means not preferred.

Let \(p_i\) denote the preference represented by the queried item, and let \(p_i^\star\) denote the sample's designated original ground-truth preference within the same facet.
We define preference popularity, \(\operatorname{pop}(p)\), as the number of distinct users whose annotations include preference \(p\) within the relevant facet.
When multiple preferences are acceptable, the popularity comparison still uses the designated original reference \(p_i^\star\).

We define the boundary-external set as
\[
\mathcal{O}
=
\{\, i \mid u_i \notin S_i,\; y_i^{*}=0 \,\}.
\]
This set contains samples where the queried item should remain outside the target user's preference boundary.
The Collapse score is computed as
\[
\begin{aligned}
\mathrm{Collapse}
&=
\frac{1}{|\mathcal{O}|}
\sum_{i\in\mathcal{O}}
\hat{y}_i\,
\mathbb{I}\!\left[
\operatorname{pop}(p_i)>
\operatorname{pop}(p_i^\star)
\right]
\\
&=
\mathbb{P}\!\left(
\hat{y}=1,\;
\operatorname{pop}(p)>\operatorname{pop}(p^\star)
\,\middle|\,
u\notin S,\; y^{*}=0
\right).
\end{aligned}
\]
Here, \(\mathbb{P}\) denotes the empirical probability over the evaluated binary preference samples.
An erroneous attribution counts as collapse only when the attributed preference is more popular than the designated reference preference.
Errors toward equally or less popular preferences are not counted as collapse, and annotated acceptable alternatives are excluded.

A higher Collapse score means that the model more often attributes more popular, non-preferred items to the target user, while a lower value indicates fewer such population-directed violations of user-specific preference boundaries.
Unlike Overall, Preference, and Profile, Collapse is not an accuracy score. Lower values are better.

\FloatBarrier
\Needspace{6\baselineskip}
\section{Bias Diagnostics and Scope of the Evaluation}
\label{app:scope_risks}

\subsection{Preference-Frequency Bias and Gender-Stratified Results}

Our analysis targets preference-frequency bias under joint multi-user learning: errors that favor common preferences at the expense of a target user's less common choices.
Reduced collapse does not by itself establish demographic fairness.
Table~\ref{tab:frequency_error_composition} reports the available accuracy and error-composition diagnostics by preference frequency.
For low-frequency preferences, accuracy rises from 28\% with full fine-tuning to 53\% with PrefMoE, while the proportion of errors classified as collapse falls from 84\% to 32\%.
The error-composition columns describe the share of collapse among incorrect predictions within a group. They are not the Collapse scores reported in the main comparison table.

\begin{table}[!htbp]
\centering
\small
\setlength{\tabcolsep}{5pt}
\renewcommand{\arraystretch}{1.12}
\caption{Available preference-frequency diagnostics. All entries are percentages. A dash denotes a value not reported in this diagnostic.}
\label{tab:frequency_error_composition}
\begin{tabular}{lcccc}
\toprule
Group & \multicolumn{2}{c}{Accuracy $\uparrow$} & \multicolumn{2}{c}{Collapse share among errors $\downarrow$} \\ \cmidrule(lr){2-3}\cmidrule(lr){4-5} & Fine-Tuning & PrefMoE & Fine-Tuning & PrefMoE \\
\midrule
High & 50 & 73 & -- & 20 \\
Mid & -- & 65 & -- & 27 \\
Low & 28 & 53 & 84 & 32 \\
\bottomrule
\end{tabular}
\end{table}

Using the available gender annotations, we report a diagnostic for 28 male and 22 female users in Table~\ref{tab:gender_diagnostic}.
PrefMoE improves accuracy and reduces reported collapse in both groups.
The observed accuracy gap is 0.0082 for full fine-tuning and 0.0061 for PrefMoE.
With only 50 users, two annotated groups, and no explicit demographic fairness objective, this comparison provides limited descriptive evidence rather than a comprehensive fairness assessment.

\begin{table}[!htbp]
\centering
\small
\setlength{\tabcolsep}{7pt}
\renewcommand{\arraystretch}{1.12}
\caption{Gender-stratified diagnostic using available user annotations (LLaVA-1.5-7B).}
\label{tab:gender_diagnostic}
\begin{tabular}{llcc}
\toprule
Group & Method & Accuracy $\uparrow$ & Collapse $\downarrow$ \\
\midrule
Male ($n=28$) & Fine-Tuning & 0.5108 & 0.3391 \\
 & PrefMoE & 0.6724 & 0.1258 \\
Female ($n=22$) & Fine-Tuning & 0.5026 & 0.3468 \\
 & PrefMoE & 0.6785 & 0.1201 \\
\bottomrule
\end{tabular}
\end{table}

\subsection{Limitations and Remaining Risks}

\paragraph{Dataset and preference scope.}
MMPB-Clean uses predefined facets and a bounded preference vocabulary.
The human-verified cross-domain set, multi-label questions, and explicit weather contexts extend the evaluation in controlled ways, but do not replace a large, naturally collected longitudinal preference dataset.
Open-ended preferences, implicit evidence, and context-dependent changes in the same user's choices remain incompletely evaluated.

\paragraph{Continual adaptation.}
The sequential experiments reveal degradation despite PrefMoE's stronger final performance.
The method does not resolve long-term forgetting, and the evaluated update horizon is insufficient to establish long-term stability.
Stale or incorrectly updated user states may continue to produce mismatched answers.

\paragraph{Privacy and fairness.}
The framework learns and retains user-specific representations, which can encode sensitive information even when the stored state is not a raw personal description.
It provides no formal privacy guarantee and does not optimize a comprehensive demographic fairness objective.
Remaining risks include demographic shortcuts from portraits or profiles, exposure of personal preferences, and feedback loops that reinforce earlier inferred preferences.
The reported reduction in preference-frequency collapse should be interpreted within this narrower scope.
\FloatBarrier
\Needspace{6\baselineskip}
\section{More Experimental Results}

\subsection{Collapse Breakdown by Preference Popularity}

Table~\ref{tab:llava_family_collapse_breakdown} reports the original popularity-stratified results across LLaVA-family backbones.
The groups are defined by how many users share the queried preference: at most four users, five to eight users, or at least nine users.
PrefMoE has lower reported collapse than the corresponding full fine-tuning baseline in each group.
Appendix~\ref{app:popularity_transitions} complements this breakdown with ground-truth-to-prediction frequency transitions and distinguishes frequency-group agreement from preference correctness.

\begin{table}[!htbp]
\centering
\caption{Collapse breakdown for LLaVA-family models across user frequency groups.}
\resizebox{0.7\linewidth}{!}{
\begin{tabular}{llcccc}
\toprule
\textbf{Method}
& \textbf{Setting}
& $\leq 4$
& 5--8
& $\geq 9$
& \textbf{Overall} \\
\midrule

LLaVA-1.5-7B
& \textit{NT}
& 0.6020 & 0.6103 & 0.6760 & 0.6228 \\

LLaVA-1.5-13B
& \textit{NT}
& 0.5750 & 0.5958 & 0.6490 & 0.6027 \\

LLaVA-OV-72B
& \textit{NT}
& 0.5140 & 0.5333 & 0.5370 & 0.5297 \\

\midrule

LLaVA-1.5-7B
& \textit{FFT}
& 0.3180 & 0.3381 & 0.3790 & 0.3425 \\

LLaVA-1.5-13B
& \textit{FFT}
& 0.2920 & 0.3056 & 0.3420 & 0.3105 \\

LLaVA-OV-72B
& \textit{FFT}
& 0.2970 & 0.3056 & 0.3580 & 0.3151 \\

\midrule

PrefMoE (LLaVA-1.5-7B)
& \textit{PEFT}
& 0.1050 & 0.1175 & 0.1570 & 0.1233 \\

PrefMoE (LLaVA-1.5-13B)
& \textit{PEFT}
& 0.1390 & 0.1330 & 0.1660 & 0.1416 \\

PrefMoE (LLaVA-OV-72B)
& \textit{PEFT}
& \textbf{0.0970} & \textbf{0.1079} & \textbf{0.1480} & \textbf{0.1142} \\

\bottomrule
\end{tabular}
}
\label{tab:llava_family_collapse_breakdown}
\end{table}

\subsection{Detailed Comparisons for Different Preference Facets}

Tables~\ref{tab:llava_family_factor_collapse} and~\ref{tab:llava_category_pref}
provide a facet-wise analysis of 0-turn preference questions across entertainment, fashion, lifestyle, shopping, and travel.
PrefMoE consistently improves preference accuracy and reduces collapse across LLaVA-family backbones, showing that its gains are not only reflected in aggregate metrics but also hold across different preference types.
For example, on LLaVA-OV-72B, PrefMoE improves the overall preference accuracy from 0.5947 to 0.7613 and reduces the overall collapse rate from 0.3151 to 0.1142 compared with FFT.

The results also reveal clear facet-level difficulty differences.
Shopping achieves particularly high preference accuracy under PrefMoE, suggesting that its cues are often more visually grounded and easier to associate with the image.
In contrast, lifestyle remains more challenging, with lower accuracy and relatively higher collapse across backbones.
This suggests that lifestyle preferences require more abstract and context-dependent personalization rather than direct visual matching.
Overall, scaling and full fine-tuning improve general performance, but they do not sufficiently preserve fine-grained preference boundaries. PrefMoE provides more consistent preference-sensitive reasoning across facets.

\begin{table}[!htbp]
\centering
\caption{Facet-wise preference collapse on the LLaVA family.}
\resizebox{\linewidth}{!}{
\begin{tabular}{llcccccc}
\toprule
\textbf{Method}
& \textbf{Setting}
& Entertainment
& Fashion
& Lifestyle
& Shopping
& Travel
& \textbf{Overall} \\
\midrule

LLaVA-1.5-7B
& \textit{NT}
& 0.6250 & 0.6458 & 0.6222 & 0.6000 & 0.6364 & 0.6228 \\

LLaVA-1.5-13B
& \textit{NT}
& 0.5833 & 0.6250 & 0.6000 & 0.5778 & 0.6061 & 0.6027 \\

LLaVA-OV-72B
& \textit{NT}
& 0.5208 & 0.5625 & 0.4889 & 0.5333 & 0.5455 & 0.5297 \\

\midrule

LLaVA-1.5-7B
& \textit{FFT}
& 0.3542 & 0.3750 & 0.3333 & 0.2889 & 0.3636 & 0.3425 \\

LLaVA-1.5-13B
& \textit{FFT}
& 0.3125 & 0.2917 & 0.2889 & 0.2667 & 0.3939 & 0.3105 \\

LLaVA-OV-72B
& \textit{FFT}
& 0.3333 & 0.3542 & 0.3111 & 0.2444 & 0.3030 & 0.3151 \\

\midrule

PrefMoE (LLaVA-1.5-7B)
& \textit{PEFT}
& 0.1458 & 0.0417 & 0.1778 & 0.0889 & 0.1212 & 0.1233 \\

PrefMoE (LLaVA-1.5-13B)
& \textit{PEFT}
& 0.1667 & \textbf{0.1042} & 0.2000 & 0.1111 & 0.0909 & 0.1416 \\

PrefMoE (LLaVA-OV-72B)
& \textit{PEFT}
& \textbf{0.1250} & 0.0625 & \textbf{0.1556} & \textbf{0.0667} & \textbf{0.0303} & \textbf{0.1142} \\

\bottomrule
\end{tabular}
}
\label{tab:llava_family_factor_collapse}

\end{table}

\begin{table}[!htbp]
\centering
\caption{Facet-wise 0-turn preference accuracy on the LLaVA family. }
\resizebox{\linewidth}{!}{
\begin{tabular}{llcccccc}
\toprule
\textbf{Method} & \textbf{Type}
& {Entertainment}
& {Travel}
& {Lifestyle}
& {Shopping}
& {Fashion}
& \textbf{Ovr.Preference} \\ 
\midrule
LLaVA-1.5-7B & \textit{NT}
& 0.3310 & 0.3440 & 0.3370 & 0.3420 & 0.3399 & 0.3387 \\
LLaVA-1.5-13B & \textit{NT}
& 0.3590 & 0.3670 & 0.3610 & 0.3710 & 0.3688 & 0.3653 \\
LLaVA-OV-72B & \textit{NT}
& 0.4740 & 0.4860 & 0.4770 & 0.4820 & 0.4812 & 0.4800 \\
\midrule
LLaVA-1.5-7B & \textit{FFT}
& 0.4370 & 0.4560 & 0.3870 & 0.4620 & 0.4626 & 0.4413 \\
LLaVA-1.5-13B & \textit{FFT}
& 0.4520 & 0.4710 & 0.4050 & 0.4860 & 0.4846 & 0.4600 \\
LLaVA-OV-72B & \textit{FFT}
& 0.5850 & 0.6130 & 0.5310 & 0.6310 & 0.6121 & 0.5947 \\
\midrule
PrefMoE (LLaVA-1.5-7B) & \textit{PEFT}
& 0.6120 & 0.6522 & 0.5860 & 0.8420 & 0.6810 & 0.6733 \\
PrefMoE (LLaVA-1.5-13B) & \textit{PEFT}
& 0.6230 & 0.6557 & 0.5920 & 0.8310 & \textbf{0.7040} & 0.6800 \\
PrefMoE (LLaVA-OV-72B) & \textit{PEFT}
& \textbf{0.7800} & \textbf{0.8594} & \textbf{0.6040} & \textbf{0.8580} & 0.6990 & \textbf{0.7613} \\
\bottomrule
\end{tabular}
}
\label{tab:llava_category_pref}
\end{table}

\subsection{Computational Cost}

\begin{table}[!htbp] 
\centering
\caption{Computational cost comparison. GPU-hours are computed as the number of GPUs multiplied by wall-clock time.}
\label{tab:computational_cost}
\resizebox{0.8\linewidth}{!}{
\begin{tabular}{lccc}
\toprule
\textbf{Method} 
& \textbf{Train Cost} 
& \textbf{Eval Cost} 
& \textbf{Peak Memory} \\
\midrule

LLaVA-1.5-7B
& -- 
& 0.5--1 GPU-hours
& 15--20 GB/GPU \\

\midrule

Yo'LLaVA-7B
& 5--10 GPU-hours 
& 1--2 GPU-hours
& 28--34 GB/GPU \\

LOVA3-7B
& 6--12 GPU-hours
& 1--2 GPU-hours
& 28--34 GB/GPU \\

TG-LLaVA-7B
& 6--12 GPU-hours 
& 1--2 GPU-hours
& 28--34 GB/GPU \\

\midrule

PrefMoE (LLaVA-1.5-7B)  
& 8--16 GPU-hours   
& 1--2 GPU-hours     
& 28--34 GB/GPU \\

PrefMoE (LLaVA-1.5-13B) 
& 24--32 GPU-hours   
& 2--3 GPU-hours    
& 34--39 GB/GPU \\

PrefMoE (LLaVA-OV-72B)   
& 450--600 GPU-hours  
& 40--80 GPU-hours  
& 38--40 GB/GPU \\

\bottomrule
\end{tabular}
}
\end{table}

Table~\ref{tab:computational_cost} reports the computational cost of different methods.
The no-training LLaVA-1.5-7B baseline only requires inference, resulting in the lowest cost and memory usage.
Existing personalized VQA methods introduce additional training overhead, but their costs remain in a similar range under 7B backbones.
Compared with these methods, PrefMoE has a slightly higher training cost on LLaVA-1.5-7B due to its preference-aware modules and auxiliary objectives, while maintaining comparable evaluation cost and peak memory.
When scaling to larger backbones, the cost increases mainly with model size: PrefMoE on LLaVA-1.5-13B requires more training resources, and the LLaVA-OV-72B variant is substantially more expensive due to the larger backbone.
Overall, PrefMoE introduces moderate overhead at the 7B scale, while the major cost increase comes from backbone scaling rather than the personalization mechanism itself.

\FloatBarrier
\Needspace{6\baselineskip}
\section{Additional Data and Evaluation Validation}
\label{app:additional_validation}

All experiments in this section use LLaVA-1.5-7B.
The 0-turn and 10-turn settings refer to dialogue history, as in the main experiments. They are distinct from the sequential model updates studied in Appendix~\ref{app:sequential_updates}.

\subsection{Popularity-Stratified Prediction Transitions}
\label{app:popularity_transitions}

\begin{figure}[!htbp]
\centering
\includegraphics[width=\linewidth]{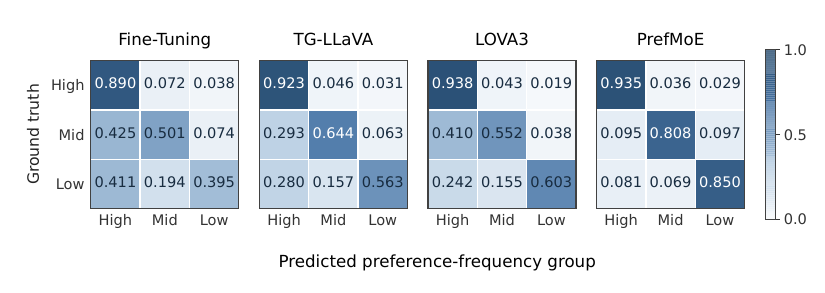}
\caption{Row-normalized transitions from ground-truth to predicted preference-frequency groups. High, Mid, and Low denote at least nine, five to eight, and at most four users, respectively. All panels share the same color scale. Each row sums to one; the diagonal denotes a matching frequency group, not exact-answer accuracy.}
\label{fig:popularity_transitions}
\end{figure}

We group preferences by their frequency across users, following the popularity thresholds used in the main appendix: High denotes at least nine users, Mid denotes five to eight users, and Low denotes at most four users.
Figure~\ref{fig:popularity_transitions} reports row-normalized transitions from the ground-truth preference group to the predicted preference group.
These coarse groups describe prediction direction: the diagonal indicates agreement in frequency group and does not necessarily indicate a correct preference prediction.
Consequently, the coarse transition matrix and the preference-level collapse score should not be treated as interchangeable measures.

PrefMoE reduces the Mid-to-High transition from 0.293 for TG-LLaVA to 0.095, and the Low-to-High transition from 0.280 to 0.081.
TG-LLaVA and LOVA3 have similar overall answer accuracies (0.5413 and 0.5329), but different transition patterns and reported collapse scores (0.2329 and 0.4292).
This comparison illustrates why answer accuracy alone does not describe the direction of preference errors.

\subsection{Multiple Acceptable Preferences}
\label{app:multilabel_validation}

We construct 40 questions with two human-verified acceptable preferences per question.
A prediction is correct if it matches either acceptable preference.
All acceptable alternatives are excluded from collapse errors before examining the direction of the remaining errors.
Table~\ref{tab:multilabel_validation} shows that PrefMoE retains the highest accuracy and lowest reported collapse among the compared methods in both dialogue settings.
The result supports the same relative ordering when more than one answer is acceptable, within this small diagnostic set.

\begin{table}[!htbp]
\centering
\small
\setlength{\tabcolsep}{6pt}
\renewcommand{\arraystretch}{1.12}
\caption{Multi-label diagnostic with 40 questions and two acceptable preferences per question.}
\label{tab:multilabel_validation}
\begin{tabular}{lcccc}
\toprule
Method & \multicolumn{2}{c}{0-turn} & \multicolumn{2}{c}{10-turn} \\ \cmidrule(lr){2-3}\cmidrule(lr){4-5} & Accuracy $\uparrow$ & Collapse $\downarrow$ & Accuracy $\uparrow$ & Collapse $\downarrow$ \\
\midrule
Fine-Tuning & 0.400 & 0.425 & 0.325 & 0.475 \\
TG-LLaVA & 0.475 & 0.325 & 0.425 & 0.375 \\
PrefMoE & 0.575 & 0.175 & 0.500 & 0.250 \\
\bottomrule
\end{tabular}
\end{table}

\subsection{Human Audit of Collapse Errors}
\label{app:collapse_human_audit}

We manually audit 20 stratified prediction errors, including both metric-positive collapse cases and metric-negative ordinary errors.
Annotators examine whether a prediction follows the population Top-1 preference or instead constitutes a plausible or non-dominant error.
The automatic classification agrees with the human judgments in all 20 audited cases.
This finding supports the interpretation of the selected cases. The small, stratified Top-1 audit does not establish a population-wide agreement rate or validate every possible form of preference shift.

\subsection{Explicit Context Conditioning}
\label{app:context_validation}

We additionally evaluate preference queries with sunny, rainy, and cloudy contexts.
Table~\ref{tab:context_validation} reports the 0-turn results.
PrefMoE obtains the highest answer accuracies and lowest reported collapse among the compared methods.
This controlled experiment tests explicit contextual conditioning, but it does not establish that the model can infer arbitrary changes in a user's preferences from naturally evolving context.

\begin{table}[!htbp]
\centering
\small
\setlength{\tabcolsep}{5pt}
\renewcommand{\arraystretch}{1.12}
\caption{Preference queries conditioned on explicit weather contexts (0-turn).}
\label{tab:context_validation}
\begin{tabular}{lcccc}
\toprule
Method & Overall $\uparrow$ & Preference $\uparrow$ & Profile $\uparrow$ & Collapse $\downarrow$ \\
\midrule
Fine-Tuning & 0.4921 & 0.4236 & 0.5289 & 0.3714 \\
Yo'LLaVA & 0.4794 & 0.4618 & 0.4889 & 0.2348 \\
TG-LLaVA & 0.5155 & 0.5524 & 0.4957 & 0.2586 \\
LOVA3 & 0.5079 & 0.5417 & 0.4898 & 0.4568 \\
PrefMoE & 0.6542 & 0.6487 & 0.6571 & 0.1519 \\
\bottomrule
\end{tabular}
\end{table}

\FloatBarrier
\Needspace{6\baselineskip}
\section{Sequential Personalization Updates}
\label{app:sequential_updates}

We study changes in personalized information separately from the dialogue-history setting.
All results in this section use LLaVA-1.5-7B.

\subsection{Two-Round Preference and Profile Updates}
\label{app:two_round_updates}

\begin{figure}[!htbp]
\centering
\includegraphics[width=\linewidth]{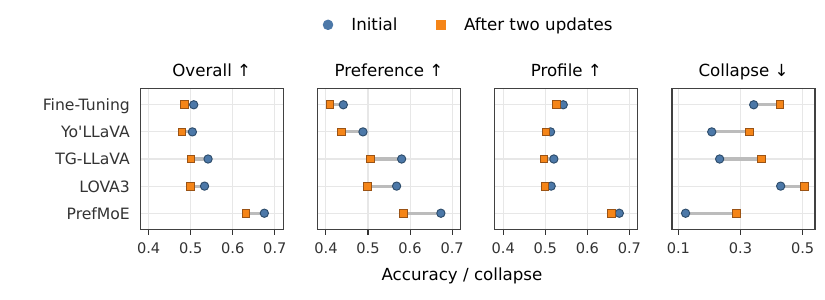}
\caption{Scores before and after two sequential rounds of preference changes. Circles show initial scores and squares show final scores; each segment connects the same method before and after updating. Overall, Preference, and Profile denote accuracies (higher-is-better); Collapse is lower-is-better.}
\label{fig:dynamic_preference}
\end{figure}

\begin{figure}[H]
\centering
\includegraphics[width=\linewidth]{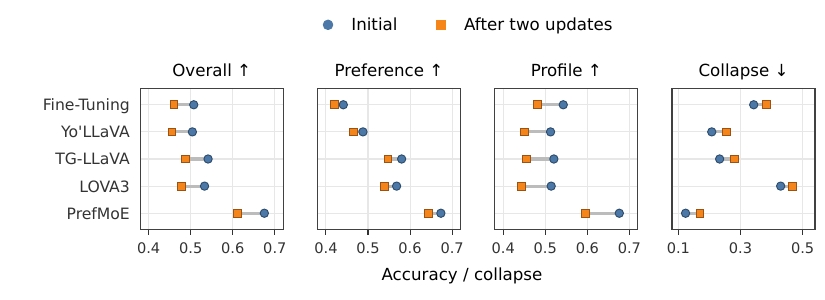}
\caption{Scores before and after two sequential rounds of profile changes, using the same layout and axis ranges as Figure~\ref{fig:dynamic_preference}. Circles show initial scores and squares show final scores. Accuracies are higher-is-better and collapse is lower-is-better.}
\label{fig:dynamic_profile}
\end{figure}

We simulate two sequential update rounds, each modifying the preference information or the profile information of 30\% of users.
The preference-change and profile-change experiments are evaluated separately.
In the profile-update experiment, PrefMoE re-encodes the updated textual profile and updates the corresponding user-specific factors, while leaving the backbone and unaffected users' representations unchanged.
Figures~\ref{fig:dynamic_preference} and~\ref{fig:dynamic_profile} compare the initial and final scores, with connecting segments showing the changes for each method.

PrefMoE retains the highest final accuracies and lowest reported collapse in both experiments.
However, its overall accuracy decreases by 0.0437 with preference changes and by 0.0634 with profile changes, compared with decreases of 0.0410 and 0.0539 for TG-LLaVA.
Its stronger final performance therefore should not be interpreted as evidence that it degrades less during updating.

\subsection{Ten Sequential Preference Updates}
\label{app:ten_updates}

We divide the users into ten groups and construct ten sequential update tasks.
At task $t$, only group $t$ undergoes a preference change, while other groups' preference states remain unchanged in that task.
Each update uses the newly changed user data without revisiting data from previous updates.
We evaluate preference accuracy using a consistent evaluation protocol across stages and report each group's trajectory from its update onward.
Unreported entries before a group's update are omitted from the plot.

\begin{figure}[H]
\centering
\includegraphics[width=\linewidth]{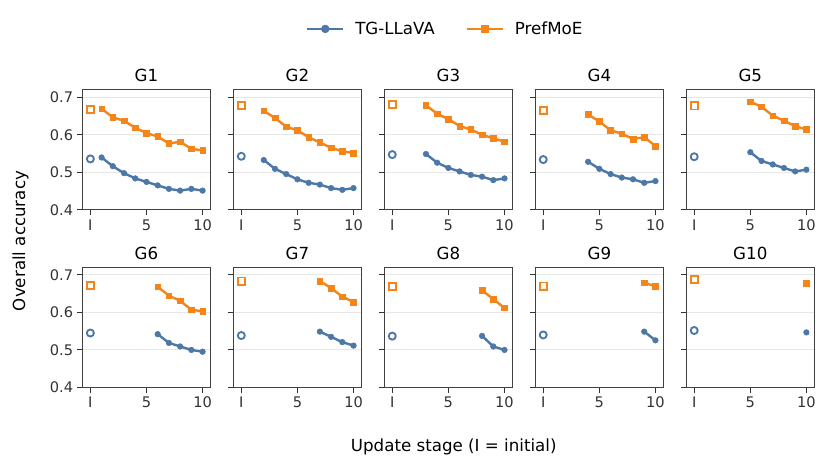}
\caption{Overall accuracy across ten sequential updates, with one panel per user group. TG-LLaVA and PrefMoE are evaluated under the same protocol. Open markers at I show initial scores; filled markers show reported update-stage scores. Group $t$ changes at update $t$, and unreported pre-update entries are omitted without interpolation. All panels share the same axes.}
\label{fig:continual_groups}
\end{figure}

Table~\ref{tab:ten_update_summary} summarizes the initial and final preference accuracies by taking an unweighted mean across the ten reported groups.
Figure~\ref{fig:continual_groups} shows every reported group trajectory alongside its initial score.
The summary drop is the mean initial preference accuracy minus the mean accuracy after update ten. Positive values indicate a decline.

After ten preference updates, PrefMoE still achieves higher
overall accuracy than TG-LLaVA (0.6065 vs. 0.4956), despite
a larger absolute decline (0.0686 vs. 0.0457).
This difference may partly reflect their unequal starting
performance: TG-LLaVA begins with substantially lower accuracy
(0.5413 vs. 0.6751), which may contribute to its smaller
absolute decline over the limited update horizon.
Nevertheless, its smaller decline does not translate into
better final performance, and PrefMoE maintains its advantage
at every reported group--update pair.

\begin{table}[H]
\centering
\small
\setlength{\tabcolsep}{7pt}
\renewcommand{\arraystretch}{1.12}
\caption{Ten-update accuracy, averaged equally across the ten user groups. These summary values are computed from the group trajectories below.}
\label{tab:ten_update_summary}
\begin{tabular}{lccc}
\toprule
Method & Initial & After update 10 & Initial-to-final drop \\
\midrule
TG-LLaVA & 0.5413 & 0.4956 & 0.0457 \\
PrefMoE & 0.6751 & 0.6065 & 0.0686\\
\bottomrule
\end{tabular}
\end{table}

\end{document}